\pdfoutput=1
\documentclass[11pt]{article}

\usepackage[final]{acl}

\usepackage[T1]{fontenc}
\usepackage[utf8]{inputenc}

\usepackage{times}
\usepackage{latexsym}

\usepackage[T1]{fontenc}
\usepackage{graphicx} % Required for inserting images
\usepackage{longtable}
\usepackage{geometry}

\PassOptionsToPackage{table,xcdraw}{xcolor}
\usepackage{booktabs}

\usepackage{inconsolata}
\usepackage{amsmath}
\usepackage{booktabs}
\usepackage{multirow}
\usepackage{graphicx}
\usepackage{caption}
\usepackage{subcaption}
\usepackage{amsfonts}
\usepackage{algorithm}
\usepackage{algorithmic}
\usepackage{amssymb}
\usepackage{pifont}
\usepackage{xcolor}
\usepackage{tabularx}
\usepackage{bbding}
\usepackage{utfsym}
\usepackage{hyperref}
\usepackage{enumitem}
\usepackage{setspace}
\usepackage{courier} % light font weight
\usepackage{url}            % simple URL typesetting
\usepackage{nicefrac}       % compact symbols for 1/2, etc.
\usepackage[final,nopatch=footnote]{microtype}
\usepackage[edges]{forest}
\usepackage{makecell}

\definecolor{hidden-red}{RGB}{205, 44, 36}
\definecolor{hidden-blue}{RGB}{194,232,247}
\definecolor{hidden-orange}{RGB}{243,202,120}
\definecolor{hidden-green}{RGB}{34,139,34}
\definecolor{hidden-pink}{RGB}{255,245,247}
\definecolor{hidden-black}{RGB}{20,68,106}

\newcommand{\eg}{E.g.,}
\newcommand{\highlightcheck}{{\raisebox{-0.2ex}{\scalebox{1.3}{\textbf{$\checkmark$}}}}}

\title{From Hypothesis to Publication: A Comprehensive Survey of AI-Driven Research Support Systems}

\author{
    \textbf{Zekun Zhou\textsuperscript{1}},
    \textbf{Xiaocheng Feng\textsuperscript{1,2}\thanks{Co-Corresponding Author}},
    \textbf{Lei Huang\textsuperscript{1}},
    \textbf{Xiachong Feng\textsuperscript{3}\footnotemark[1]},
    \\
    \textbf{Ziyun Song\textsuperscript{1}},
    \textbf{Ruihan Chen\textsuperscript{1}},
    \textbf{Liang Zhao\textsuperscript{1}},
    \textbf{Weitao Ma\textsuperscript{1}},
    \textbf{Yuxuan Gu\textsuperscript{1}},
    \\
    \textbf{Baoxin Wang\textsuperscript{4}},
    \textbf{Dayong Wu\textsuperscript{4}},
    \textbf{Guoping Hu\textsuperscript{4}},
    \textbf{Ting Liu\textsuperscript{1}},    
    \textbf{Bing Qin\textsuperscript{1,2}}
    \\
    \textsuperscript{1} Harbin Institute of Technology \quad
    \textsuperscript{2} Peng Cheng Laboratory \\
    \textsuperscript{3} The University of Hong Kong \quad
    \textsuperscript{4} iFLYTEK Research \\
    \texttt{\{zkzhou,xcfeng,lhuang,yxgu,qinb\}@ir.hit.edu.cn}
    \\
    \texttt{fengxc@hku.hk} \quad \texttt{\{bxwang2,dywu2,gphu\}@iflytek.com}
}

\begin{document}

\maketitle

\begin{abstract}
Research is a fundamental process driving the advancement of human civilization, yet it demands substantial time and effort from researchers. In recent years, the rapid development of artificial intelligence (AI) technologies has inspired researchers to explore how AI can accelerate and enhance research. To monitor relevant advancements, this paper presents a systematic review of the progress in this domain. Specifically, we organize the relevant studies into three main categories: hypothesis formulation, hypothesis validation, and manuscript publication. Hypothesis formulation involves knowledge synthesis and hypothesis generation. Hypothesis validation includes the verification of scientific claims, theorem proving, and experiment validation. Manuscript publication encompasses manuscript writing and the peer review process. Furthermore, we identify and discuss the current challenges faced in these areas, as well as potential future directions for research. Finally, we also offer a comprehensive overview of existing benchmarks and tools across various domains that support the integration of AI into the research process. We hope this paper serves as an introduction for beginners and fosters future research. Resources have been made publicly
available\footnote{\url{https://github.com/zkzhou126/AI-for-Research}}.

\end{abstract}
\section{Introduction} \label{sec:introduction}

Research is creative and systematic work aimed at expanding knowledge and driving civilization's development~\citep{eurostat2018measurement}. Researchers typically identify a topic, review relevant literature, synthesize existing knowledge, and formulate hypothesis, which are validated through theoretical and experimental methods. Findings are then documented in manuscripts that undergo peer review before publication~\citep{benos2007ups,2311JamesBoykoAn}. However, this process is resource-intensive, requiring specialized expertise and posing entry barriers for researchers~\citep{10BlaxterHow}.

In recent years, artificial intelligence (AI) technologies, represented by large language models (LLMs), have experienced rapid development ~\citep{20BrownLanguage,2303OpenAIGPT4,2407AbhimanyuLlama3,2412AnYangQwen2.5,2412DeepSeekAIDeepSeekV3,2501deepseekAIDeepseek-r1}. These models exhibit exceptional capabilities in text understanding, reasoning, and generation~\citep{23RylanSchaefferAlice}. In this context, AI is increasingly involving the entire research pipeline~\citep{2024MesseriArtificial}, sparking extensive discussion about its implications for research ~\citep{22HutsonCould,2303NigelLWilliamsAlgorithmic,2304MeredithScientists,2306BenediktFecherFriend}. Moreover, following the release of ChatGPT, approximately 20\% of academic papers and peer-reviewed texts in certain fields have been modified by LLMs~\citep{2403WeixinLiangMonitoring,2404WeixinLiangMapping}. A study also reveals that 81\% of researchers integrate LLMs into their workflows~\citep{2411ZhehuiLiaoLLMs}. 

% \begin{figure}[t]   
%     \centering
%         \includegraphics[clip]{figures/overview.pdf}
%         \caption{\label{figure:paper_overview}
%        Overview of AI for research. The framework consists of three stages: hypothesis formulation, hypothesis validation, and manuscript publication. In the hypothesis formulation stage, knowledge integration leads to the proposal of an initial hypothesis after a topic is identified. The hypothesis validation stage involves verifying the hypothesis from three perspectives to ensure its correctness and validity. Finally, the manuscript publication stage focuses on drafting and publishing the validated hypothesis.
%        }
% % \vspace{-5mm}
% \end{figure}

\begin{figure}[t]
    \centering
    % Make the image width equal to the column width
    \includegraphics[width=\columnwidth, clip]{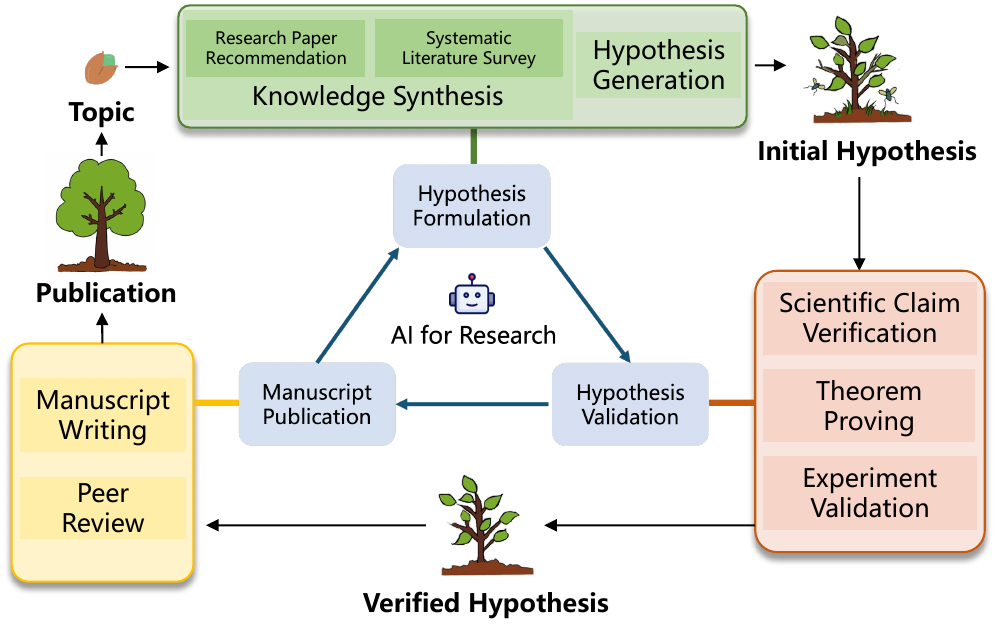}
    \caption{\label{figure:paper_overview}
    Overview of AI for research. The framework consists of three stages: hypothesis formulation, hypothesis validation, and manuscript publication. In the hypothesis formulation stage, knowledge integration leads to the proposal of an initial hypothesis after a topic is identified. The hypothesis validation stage involves verifying the hypothesis from three perspectives to ensure its correctness and validity. Finally, the manuscript publication stage focuses on drafting and publishing the validated hypothesis.
    }
% \vspace{-5mm} % Uncomment if you need to adjust vertical spacing
\end{figure}

As the application of AI in research attracts increasing attention, a significant body of related studies has begun to emerge. To systematically synthesize existing research, we present comprehensive survey that emulates human researchers by using the research process as an organizing framework. Specifically, as depicted in Figure~\ref{figure:paper_overview}, the research process is divided into three key stages: (1) Hypothesis Formulation, involving knowledge synthesis and 
hypothesis generation; (2) Hypothesis Validation,  encompassing scientific claim verification, theorem proving, and experiment validation; (3) Manuscript Publication, which focuses on academic publications and is further divided into manuscript writing and peer review. 

\textbf{Comparing with Existing Surveys} Although~\citet{25LuoZimingLLM4SR} reviews the application of AI in research, it predominantly focuses on LLMs, while neglecting the knowledge synthesis that precedes hypothesis generation and the theoretical validation of hypothesis. Other surveys concentrate on more specific areas, such as paper recommendation~\citep{16beelpaper, 19XiaomeiBaiScientific, 2201ChristinKatharinaKreutzScientific}, scientific literature review~\citep{22NoufIbrahimAltmamiAutomatic}, hypothesis generation~\citep{kulkarni2025scientific}, scientific claim verification~\citep{2305JurajVladikaScientific, 2408AlphaeusDmonteClaim}, theorem proving~\citep{2404ZhaoyuLiA}, manuscript writing~\citep{2404XiangciLiRelated}, and peer review ~\citep{2111JialiangLinAutomated, 23KayvanKoushaArtificial}. Additionally, certain surveys emphasize the application of AI in scientific domains~\citep{2310YizhenZhengLarge,2406YuZhangA, gridach2025agentic}. 

\textbf{Contributions} Our contributions can be summarized as follows: (1) We align the relevant fields with the research process of human researchers, systematically integrating and extending these aspects while primarily focusing on the research process itself. (2) We introduce a meticulous taxonomy (shown in Figure~\ref{fig:taxonomy}). (3) We provide a summary of tools that can assist in the research process. (4) We discuss new frontiers, outline their challenges, and shed light on future research.

\tikzstyle{my-box}=[
    rectangle,
    draw=hidden-black,
    rounded corners,
    text opacity=1,
    minimum height=1.5em,
    minimum width=5em,
    inner sep=2pt,
    align=center,
    fill opacity=.5,
]
\tikzstyle{leaf}=[
    my-box, 
    minimum height=1.5em,
    % fill=hidden-orange!60, 
    fill=hidden-blue!90, 
    text=black,
    align=left,
    font=\normalsize,
    inner xsep=2pt,
    inner ysep=4pt,
]
\begin{figure*}[t]
    \vspace{-2mm}
    \centering
    \resizebox{\textwidth}{!}{
        \begin{forest}
            forked edges,
            for tree={
                child anchor=west,
                parent anchor=east,
                grow'=east,
                anchor=west,
                base=left,
                font=\large,
                rectangle,
                draw=hidden-black,
                rounded corners,
                align=left,
                minimum width=4em,
                edge+={darkgray, line width=1pt},
                s sep=3pt,
                inner xsep=2pt,
                inner ysep=3pt,
                line width=0.8pt,
                ver/.style={rotate=90, child anchor=north, parent anchor=south, anchor=center},
            },
            where level=1{text width=8em,font=\normalsize,}{},
            where level=2{text width=13em,font=\normalsize,}{},
            where level=3{text width=11.5em,font=\normalsize,}{},
            where level=4{text width=13em,font=\normalsize,}{},
            [
                A survey of AI for Research, ver
                [
                    Hypothesis\\
                    Formulation ~(\S\ref{sec:HypothesisFormulation})
                    [
                        Knowledge Synthesis ~(\S\ref{sec:HypothesisFormulation_KnowledgeSynthesis})
                        [
                            Research Paper \\
                            Recommendation ~(\S\ref{sec:HypothesisFormulation_KnowledgeSynthesis_RPR})
                            [
                                \eg~
                                ComLittee~\cite{2302HyeonsuB.KangComLittee} {,}
                                PaperWeaver~\cite{2403YoonjooLeePaperWeaver}
                                ,leaf, text width=34.5em
                            ]
                        ]
                        [
                            Systematic \\
                            Literature Review ~(\S\ref{sec:HypothesisFormulation_KnowledgeSynthesis_SLR})
                            [
                                \eg~
                                 AutoSurvey ~\cite{2406YidongWangAutoSurvey} {,}
                                 OpenScholar ~\cite{2411AkariAsaiOpenScholar}
                                ,leaf, text width=34.5em
                            ]
                        ]
                    ]
                    [
                        Hypotheses Generation ~(\S\ref{sec:HypothesisFormulation_HypothesesGeneration})
                        [
                            \eg~
                             SciMON ~\cite{2305QingyunWangSciMON} {,}
                             MOOSE ~\cite{2309ZonglinYangLarge} {,}
                             Dolphin ~\cite{2501YuanJiakangDolphin}
                            ,leaf, text width=47.6em
                        ]
                    ]
                ]
                [
                    Hypothesis\\
                    Validation  ~(\S\ref{sec:HypothesisValidation})
                    [
                        Scientific \\
                        Claim Verification ~(\S\ref{sec:HypothesisValidation_ScientificClaimVerification})
                        [
                            \eg~
                             FactKG ~\cite{2305JihoKimFactKG} {,}                         
                             SFAVEL~\cite{2309AdrinBazagaUnsupervised} {,}
                             ClaimVer~\cite{2403PreetamPrabhuSrikarDammuClaimVer}
                            ,leaf, text width=47.6em
                        ]
                    ]
                    [
                        Theorem Proving ~(\S\ref{sec:HypothesisValidation_TheoremProving})
                        [
                            \eg~
                             Dt-solver ~\cite{23DT-SolverHaimingWang} {,}
                             LEGO-Prover ~\cite{2310HaimingWangLEGO-Prover} {,}
                             DeepSeek-Prover ~\cite{2410HuajianXinDeepSeek-Prover}
                            ,leaf, text width=47.6em
                        ]
                    ]
                    [
                        Experiment Validation ~(\S\ref{sec:HypothesisValidation_ExperimentVerification})
                        [
                            \eg~
                            SANDBOX ~\cite{2305RuiboLiuTraining} {,}
                            CRISPR-GPT ~\cite{2404KaixuanHuangCRISPRGPT} {,}
                            MLR-Copilot ~\cite{2408RuochenLiMLRCopilot}
                            ,leaf, text width=47.6em
                        ]
                    ]
                ]
                [
                    Manuscript\\
                    Publication ~(\S\ref{sec:ManuscriptPublication})
                    [
                        Manuscript Writing ~(\S\ref{sec:ManuscriptPublication_ManuscriptWriting})
                        [
                            \eg~
                            Scilit ~\cite{2306NianlongGuSciLit} {,}
                            UR3WG ~\cite{23ZhengliangShiTowards} {,}
                            STEP-BACK ~\cite{2406XiangruTangStep}
                            ,leaf, text width=47.6em
                        ]
                    ]
                    [
                        Peer Review ~(\S\ref{sec:ManuscriptPublication_PeerReview})
                        [
                            \eg~
                            SWIF2T ~\cite{2405EricChamounAutomated} {,}
                            GLIMPSE ~\cite{2406MaximeDarrinGLIMPSE} {,}
                            MetaWriter ~\cite{24LuSunMetaWriter}
                            ,leaf, text width=47.6em
                        ]
                    ]
                ]
            ]
        \end{forest}
    }
    %\vspace{-4mm}
    \caption{Taxonomy of Hypothesis
Formulation, Hypothesis Validation and Manuscript
Publication (This is a simplified version, full version in Figure~\ref{fig:taxonomy_full}).}
    \label{fig:taxonomy}
    % \vspace{-3mm}
\end{figure*}
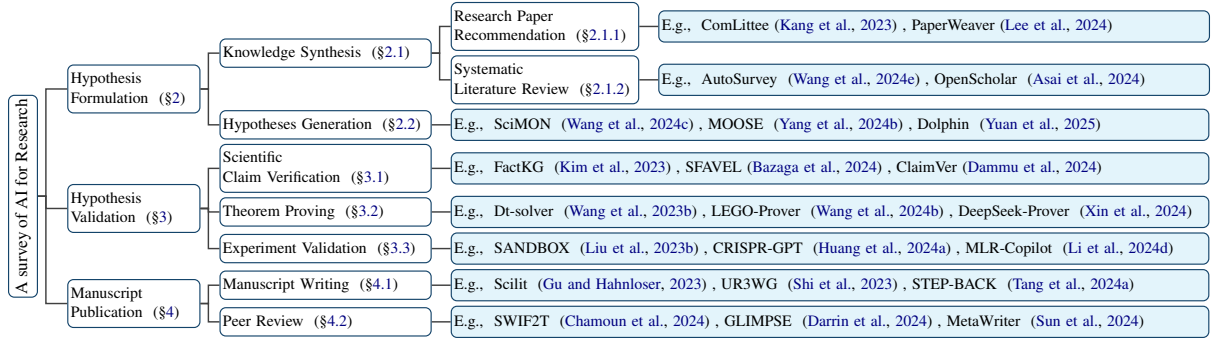

\textbf{Survey Organization} We first elaborate hypothesis formulation~(\S\ref{sec:HypothesisFormulation}), followed by hypothesis validation~(\S\ref{sec:HypothesisValidation}) and manuscript publication~(\S\ref{sec:ManuscriptPublication}). Additionally, we present benchmarks~(\S\ref{sec:Benchmarks}), and tools~(\S\ref{sec:Tools}) that utilized in research. Finally, we outline challenges as well as future directions~(\S\ref{sec:Challenges}), and discusssion about relevant ethical considerations~(\S\ref{sec:Ethical Considerations}). In the Appendix, we provide further discussion on open questions~(\S\ref{sec:discussion}), challenges faced in different domains~(\S\ref{sec:appendixchallenges}), and a comparison of capabilities among different methods~(\S\ref{sec:Ability Comparison}).

\section{Hypothesis Formulation} \label{sec:HypothesisFormulation}
\begin{figure}[t]   
    \centering
        \includegraphics[clip, width=\linewidth]{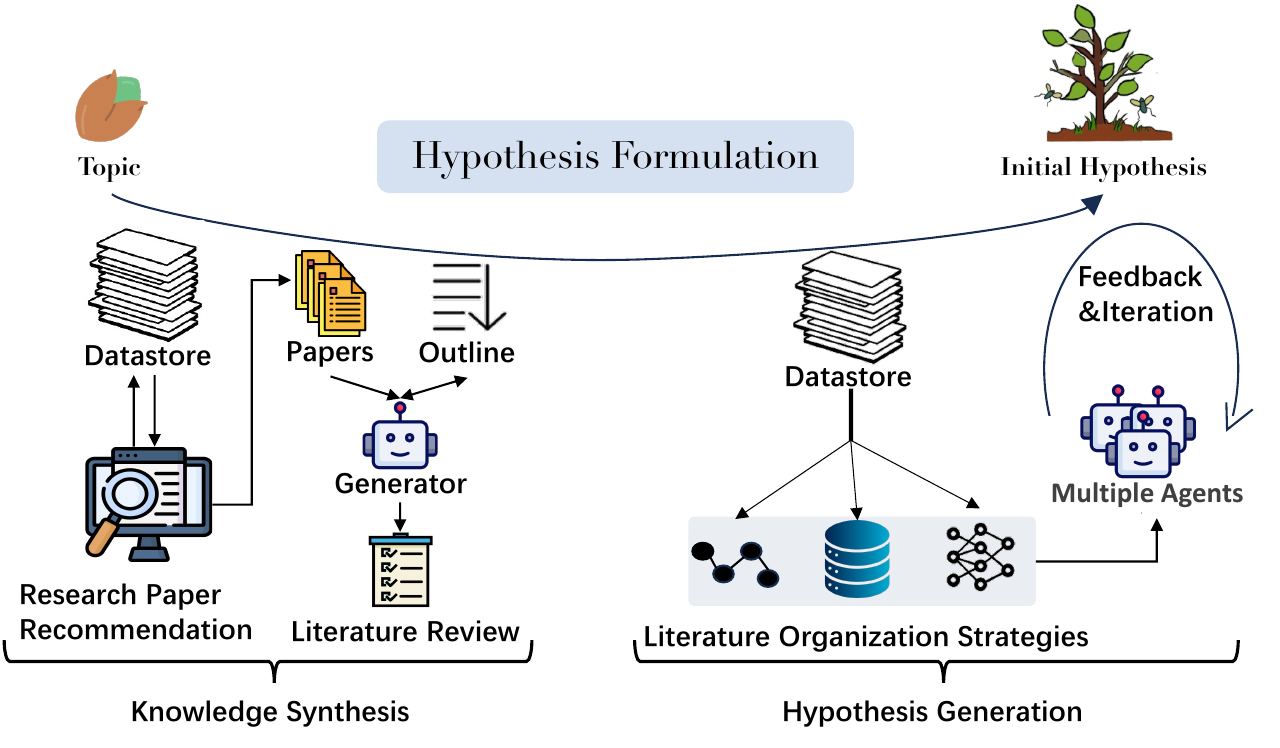}
        \caption{\label{figure:HypothesisFormulation}
       This figure illustrates the hypothesis formulation process, consisting of two stages: knowledge synthesis and hypothesis generation, which together produce an initial hypothesis related to a specific topic.
        }

% \vspace{-5mm}
\end{figure}

This stage centers on the process of hypothesis formulation. As illustrated in Figure~\ref{figure:HypothesisFormulation}, it commences with developing a comprehensive understanding of the domain, followed by identifying a specific aspect and generating pertinent hypothesis. This section is further structured into two key components: Knowledge Synthesis and Hypothesis Generation.

\subsection{Knowledge Synthesis} \label{sec:HypothesisFormulation_KnowledgeSynthesis}
Knowledge synthesis constitutes the foundational step in the research process. During this phase, researchers are required to identify and critically evaluate existing literature to establish a thorough understanding of the field. This step is pivotal for uncovering new research directions, refining methodologies, and supporting evidence-based decision-making~\citep{2411AkariAsaiOpenScholar}. In this section, the process of knowledge synthesis is divided into two modules: Research Paper Recommendation and Systematic Literature Review.

\subsubsection{Research Paper Recommendation} \label{sec:HypothesisFormulation_KnowledgeSynthesis_RPR}
Research paper recommendation (RPR) identifies and recommends novel and seminal articles aligned with researchers' interests. Prior studies have shown that recommendation systems outperform keyword-based search engines in terms of efficiency and reliability when extracting valuable insights from large-scale datasets~\citep{19XiaomeiBaiScientific}. Existing methodologies are broadly categorized into four paradigms: content-based filtering, collaborative filtering, graph-based approaches, and hybrid systems~\citep{16beelpaper,1901ZhiLiA,19XiaomeiBaiScientific,20shahidinsights}. Recent advancements propose multi-dimensional classification frameworks based on data source utilization ~\citep{2201ChristinKatharinaKreutzScientific}.

Recent trends in research suggest a decline in publication volumes related to RPR ~\citep{23RituSharmaAn}, alongside an increasing focus on user-centric optimizations. Existing studies emphasize the limitations of traditional paper-centric interaction models and advocate for more effective utilization of author relationship graphs ~\citep{2302HyeonsuB.KangComLittee}. Multi-stage recommendation architectures that integrate diverse methodologies have been shown to achieve superior performance ~\citep{2410IratxePinedoArZiGo,24VaiosStergiopoulosAn}. Visualization techniques that link recommended papers to users' publication histories via knowledge graphs ~\citep{22HyeonsuBKangFrom} and LLMs-powered comparative analysis frameworks ~\citep{2403YoonjooLeePaperWeaver} represent emerging directions for enhancing interpretability and contextual relevance.

\subsubsection{Systematic Literature Review} \label{sec:HypothesisFormulation_KnowledgeSynthesis_SLR}
Systematic literature review (SLR) constitutes a rigorous and structured methodology for evaluating and integrating prior research on a specific topic ~\citep{02JaneWebsterAnalyzing, 2304KunZhuHierarchical,2402FranciscoBolaArtificial}. In contrast to single-document summaries~\citep{elhadad2005customization}, SLR entails synthesizing information across multiple related scientific documents~\citep{22NoufIbrahimAltmamiAutomatic}. SLR can further be divided into two stages: outline generation and full-text generation~\citep{2402YijiaShaoAssisting,2412agarwalllms,24BlockWhat}. 

\textbf{Outline generation}, especially structured outline generation, is highlighted by recent studies as a pivotal factor in enhancing the quality of surveys.  ~\citet{2304KunZhuHierarchical} demonstrated that hierarchical frameworks substantially enhance survey coherence. AutoSurvey ~\citep{2406YidongWangAutoSurvey} extended conventional outline generation by recommending both sub-chapter titles and detailed content descriptions, ensuring comprehensive topic coverage. Additionally, multi-level topic generation via clustering methodologies has been proposed as an effective strategy for organizing survey structures  ~\citep{2408UriKatzKnowledge}. Advanced systems such as STORM ~\citep{2402YijiaShaoAssisting} employed LLM-driven outline drafting combined with multi-agent discussion cycles to iteratively refine the generated outlines. Tree-based hierarchical architectures have gained increasing adoption in this domain. For instance, CHIME ~\citep{2407ChaoChunHsuCHIME} optimized LLM-generated hierarchies through human-AI collaboration, while HiReview ~\citep{2410HuYuntongHireview} demonstrated the efficacy of multi-layer tree representations for systematic knowledge organization.

\textbf{Full-text generation} follows the outline generation stage. AutoSurvey and ~\citet{2408YuxuanLaiInstruct} utilized LLMs with carefully designed prompts to construct comprehensive literature reviews step-by-step. PaperQA2 ~\citep{2409MichaelDSkarlinskiLanguage} introduced an iterative agent-based approach for generating reviews, while STORM employed multi-agent conversation data for this purpose. LitLLM ~\citep{2402ShubhamAgarwalLitLLM} and ~\citet{2412agarwalllms} adopted a plan-based search enhancement strategy. KGSum ~\citep{2209PanchengWangMulti-Document} integrated knowledge graph information into paper encoding and used a two-stage decoder for summary generation. Bio-SIEVE ~\citep{2308AmbroseRobinsonBio-SIEVE} and ~\citet{2404Teosusnjakautomating} fine-tuned LLMs for automatic review generation. OpenScholar ~\citep{2411AkariAsaiOpenScholar} developed a pipeline that trained a new model without relying on a dedicated survey-generation model.

\subsection{Hypothesis Generation} \label{sec:HypothesisFormulation_HypothesesGeneration}
Hypothesis generation, known as idea generation, refers to the process of coming up with new concepts, solutions, or approaches. It is the most important step in driving the progress of the entire research ~\citep{2311BiQingQiLarge}. 

Early work focused more on predicting relationships between concepts, because researchers believed that new concepts come from links with old concepts~\citep{1708SamHenryLiterature,2210MarioKrennPredicting}. As language models became more powerful~\citep{LLMsurveyzhaoxin}, researchers are beginning to focus on open-ended idea generation ~\citep{23GirotraIdeas,2409ChengleiSiCan,2409SandeepKumarCan}. Recent advancements in AI-driven hypothesis generation highlight diverse approaches to research conceptualization. For instance, MOOSE-Chem ~\citep{2410ZonglinYangMOOSE} and IdeaSynth ~\citep{2410KevinPuIdeaSynth} used LLMs to bridge inspiration-to-hypothesis transformation via interactive frameworks. The remaining research primarily falls into two areas: enhancing input data quality and improving the quality of generated hypothesis.

\textbf{Input data quality improvement} is demonstrated by \citet{2402BodhisattwaPrasadMajumder, 2410HaokunLiuLiterature}, who showed that LLMs can generate comprehensive hypothesis from existing academic data. Literature organization strategies have evolved through various methodologies, including triplet representations ~\citep{2305QingyunWangSciMON}, chain-based architectures ~\citep{2410LongLiChain}, and complex database systems ~\citep{2410WenxiaoWangSciPIP}. Knowledge graphs emerge as critical infrastructure~\citep{21HoganKnowledge}, enabling semantic relationship mapping via subgraph identification~\citep{2403MarkusJBuehlerAccelerating,2409AlirezaGhafarollahiSciAgents}. Notably, SciMuse ~\citep{2405XuemeiGuGeneration} pioneered researcher-specific hypothesis generation by constructing personalized knowledge graphs. 

\textbf{Hypothesis quality improvement} has been addressed through feedback and iteration~\citep{rabby2025iterative}, as proposed by HypoGeniC ~\citep{2404YangqiaoyuZhouHypothesis} and MOOSE ~\citep{2309ZonglinYangLarge}. Feedback mechanisms include direct responses to hypothesis ~\citep{2404JinheonBaekResearchAgent}, experimental outcome evaluations ~\citep{2405PingchuanMaLLM, 2501YuanJiakangDolphin}, comparison with the existing literature~\citep{schmidgall2025agentrxiv}, and automated peer review comments ~\citep{2408ChrisLuThe}. FunSearch~\citep{romera2024mathematical} generates solutions by iteratively combining the innovative capabilities of LLM with the verification capabilities of an evaluator. Beyond iterative feedback, collaborative efforts among researchers have also been recognized, leading to multi-agent hypothesis generation approaches ~\citep{2403HarshitNigamAcceleron, 2409AlirezaGhafarollahiSciAgents}. VIRSCI ~\citep{2410HaoyangSuTwo} further optimized this process by customizing knowledge for each agent. Additionally, the Nova framework ~\citep{2410XiangHuNova} was introduced to refine hypothesis by leveraging outputs from other research as input.

Knowledge synthesis and hypothesis generation comprise the hypothesis formulation phase. Research paper recommendation supports knowledge acquisition, while systematic literature review aids organization within knowledge synthesis. Recent advances integrate LLMs ~\citep{2401JosedelaTorreLopezArtificial} to enhance knowledge integration ~\citep{23HuangJingshanThe, 23GuptaRohunUtilization, 24KacenaMelissaAThe, 2412TangXuemeiAre}. By developing a deep understanding of a domain through knowledge synthesis, researchers can identify research directions and use hypothesis generation techniques to formulate  hypothesis. Additionally, the distinction between scientific discovery and hypothesis generation is discussed in ~\S\ref{sec:discussion}.

\section{Hypothesis Validation} \label{sec:HypothesisValidation}
\begin{figure}[t]   
    \centering
        \includegraphics[clip, width=\linewidth]{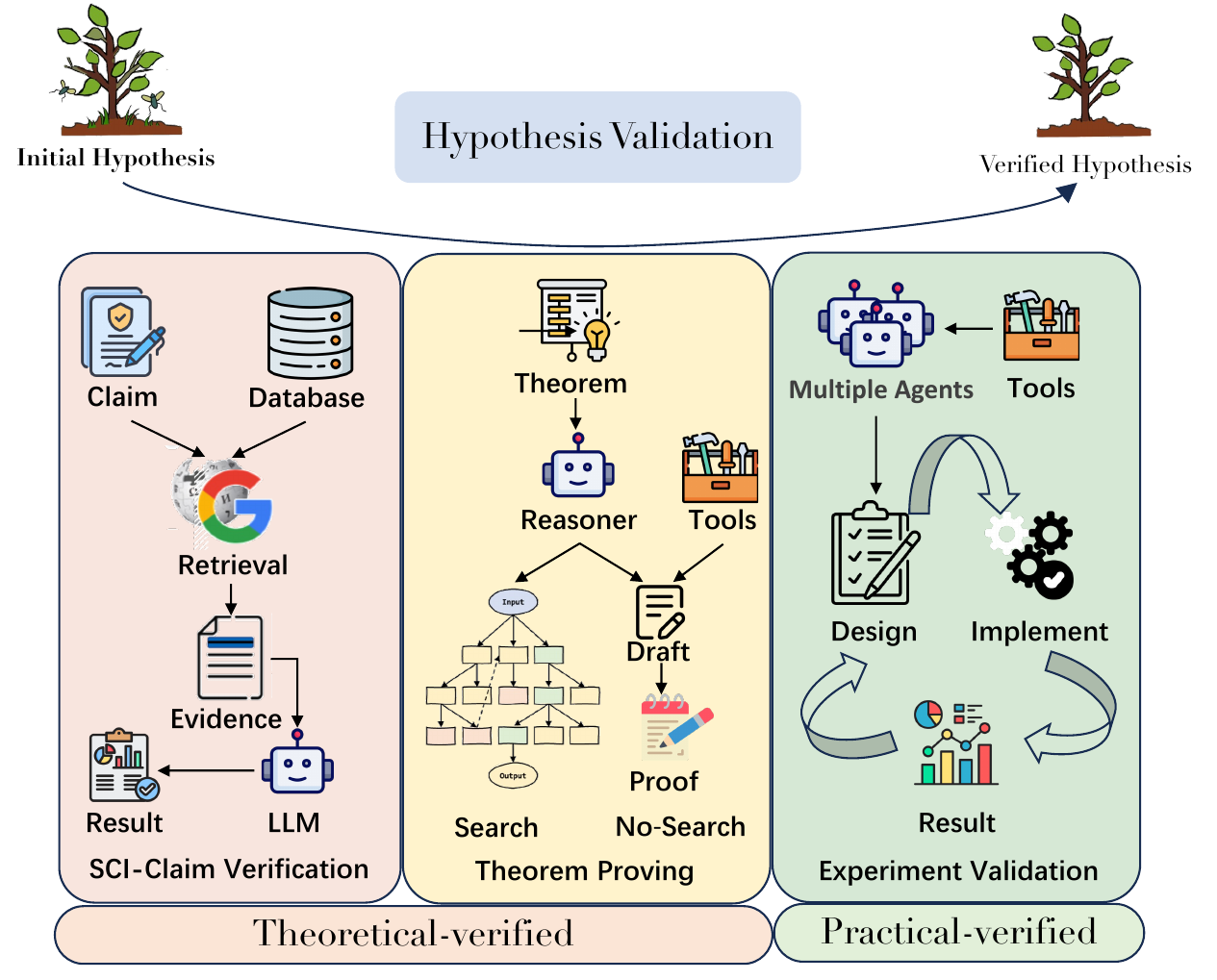}
        \caption{\label{figure:HypothesisValidation}
        This figure illustrates the various perspectives for hypothesis validation during the hypothesis validation stage. A hypothesis is typically divided into scientific claims and theorems, with SCI-claim verification (scientific claim verification) and theorem proving ensuring theoretical correctness, while experiment validation assesses practical feasibility.
        }

% \vspace{-5mm}
\end{figure}

In scientific research, any proposed hypothesis must undergo rigorous validation to establish its validity. In some studies, this process is also referred to as 'falsification'~\citep{2411ZijunLiuAIGS, 2502KexinHuangAutomated}. As illustrated in Figure ~\ref{figure:HypothesisValidation}, this section explores the application of AI in verifying scientific hypothesis through three methodological components: Scientific Claim Verification, Theorem Proving, and Experiment Validation.

\subsection{Scientific Claim Verification} \label{sec:HypothesisValidation_ScientificClaimVerification}
Scientific claim verification, also referred to as scientific fact-checking or scientific contradiction detection, aims to assess the veracity of claims related to scientific knowledge. This process assists scientists in verifying research hypothesis, discovering evidence, and advancing scientific work ~\citep{2004DavidWaddenFact, 2305JurajVladikaScientific, 2409MichaelDSkarlinskiLanguage}. Research on scientific claim verification primarily focuses on three key elements: the claim, the evidence, and the validity of the claim ~\citep{2408AlphaeusDmonteClaim}.

\paragraph{Claim} Studies have highlighted that certain claims lack supporting evidence ~\citep{2402AmelieWhrlWhat}, while others have demonstrated the ability to perform claim-evidence alignment without annotated data ~\citep{2309AdrinBazagaUnsupervised}. Additionally, methods such as HiSS ~\citep{2310XuanZhangTowards} and ProToCo ~\citep{2306FengzhuZengPrompt} proposed generating multiple claim variants to enhance verification.

\paragraph{Evidence} Existing research has explored various aspects related to evidence, including evidentiary sources ~\citep{2402JurajVladikaComparing}, retrieval configurations ~\citep{2404JurajVladikaImproving}, strategies for identifying and mitigating flawed evidence ~\citep{2210MaxGlocknerMissing, 2402AmelieWhrlUnderstanding, 2408MaxGlocknerGrounding}, and approaches for processing sentence-level ~\citep{2309LiangmingPanInvestigating} versus document-level indicators ~\citep{2112DavidWaddenMultiVerS}.

\paragraph{Verification} In the verification results generation phase, MAGIC~\citep{24WeiYuKaoMAGIC} and SERIf~\citep{2402YupengCaoCan} proposed utilizing LLMs to synthesize evidence into more comprehensive information. FactKG ~\citep{2305JihoKimFactKG} and ~\citet{2409AriefPurnamaMuharramEnhancing} structured evidence into knowledge graphs, enabling claim attribution  ~\citep{2403PreetamPrabhuSrikarDammuClaimVer, 23JinxuanWuCharacterizing}. Furthermore, ~\citet{2004PepaAtanasovaGenerating, 2108AmrithKrishnaProoFVer, 2305LiangmingPanFact-Checking, 2407IslamEldifrawiAutomated, 2410XiaochengZhangAugmenting} advocated for generating explanatory annotations alongside experimental outcomes during the verification process. Meanwhile, ~\citet{2301AnubrataDasThe, 2305EnesAltuncuaedFaCT} emphasized the critical role of domain expertise in ensuring accurate verification.

\subsection{Theorem Proving} \label{sec:HypothesisValidation_TheoremProving}
Theorem proving constitutes a subtask of logical reasoning, aimed at reinforcing the validity of the underlying theory within a hypothesis ~\citep{pease2019importance, 2303ZonglinYangLogical, 2404ZhaoyuLiA}.

Following the proposal of GPT-f ~\citep{2009StanislasPoluGenerative} to utilize generative language models for theorem proving,  researchers initially combined search algorithms with language models ~\citep{2205GuillaumeLampleHyperTree, 23DT-SolverHaimingWang}. However, a limitation in search-based approaches was later identified by ~\citet{2405HaimingWangProving}, who highlighted their tendency to explore insignificant intermediate conjectures. This led some teams to abandon search algorithms entirely. Subsequently, alternative methods emerged, such as the two-stage framework proposed by \citet{2210AlbertQiaochuJiangThe} and \citet{2407HaohanLinLean-STaR}, which prioritized informal conceptual generation before formal proof construction. Thor ~\citep{2407AlbertQiaochuJiangAdvances} introduced theorem libraries to accelerate proof generation, an approach enhanced by Logo-power ~\citep{2310HaimingWangLEGO-Prover} through dynamic libraries. Studies like Baldur ~\citep{2303EmilyFirstBaldur}, Mustard ~\citep{2303YinyaHuangMUSTARD}, and DeepSeek-Prover ~\citep{2410HuajianXinDeepSeek-Prover} demonstrated improvements via targeted data synthesis and fine-tuning, though COPRA ~\citep{24ThakurAn} questioned their generalizability and proposed an environment-agnostic alternative. Complementary strategies included theoretical decomposition into sub-goals ~\citep{2305XueliangZhaoDecomposing} and leveraging LLMs as collaborative assistants in interactive environments ~\citep{2404PeiyangSongTowards}.

\subsection{Experiment Validation} \label{sec:HypothesisValidation_ExperimentVerification}
Experiment validation involves designing and conducting experiments based on the hypothesis. The empirical validity of the hypothesis is then determined through analysis of the experimental results~\citep{2310QianHuangMLAgentBench}.

Experiment validation represents a time-consuming component of scientific research.  Recent advancements in LLMs have enhanced their ability to plan and reason about experimental tasks ~\citep{2402SubbaraoKambhampatiPosition}, prompting researchers to use these models for designing and implementing experiments ~\citep{24RuanYixiangAn}. To ensure accuracy, studies such as ~\citet{2305ShujianZhangAutoMLGPT} and ~\citet{2406SrenArltMetaDesigning} imposed input/output constraints, though this reduced generalizability. To address this, ~\citet{23DaniilABoikoAutonomous, 2304AndresMBranAugmenting, 2404KaixuanHuangCRISPRGPT} integrated tools to expand model capabilities. Full automation was achieved by ~\citet{2311BoNiMechAgents, 2410LongLiChain, 2408ChrisLuThe} through prompt-guided multi-agent collaboration. ~\citet{2303AmanMadaanSelfRefine,2501YuanJiakangDolphin} further highlighted that the integration of feedback mechanisms demonstrated potential for enhancing design quality, while ~\citet{2304LeiZhangMLCopilot,2402SiyiLiuLarge,2411ZiqiNiMatPilot} employed experimental outcomes to refine hyperparameter configurations, and ~\citet{23SzymanskiNathanJautonomousAn,2408RuochenLiMLRCopilot,2404JinheonBaekResearchAgent} leveraged agent-generated analytical insights to facilitate iterative hypothesis refinement. In contrast, social science research often uses LLMs as experimental subjects to simulate human participants ~\citep{2305RuiboLiuTraining, 2404ManningBenjaminSAutomated, 2412XinyiMouFrom}.

A hypothesis can be conceptualized as consisting of two key components: claims and theorems. Scientific claim verification and theorem proving offer theoretical validation of hypothesis through formal reasoning and logical deduction, whereas experiment validation provides comprehensive practical validation via empirical testing.

\section{Manuscript Publication} \label{sec:ManuscriptPublication}
\begin{figure}[t]   
    \centering
        \includegraphics[clip, width=\linewidth]{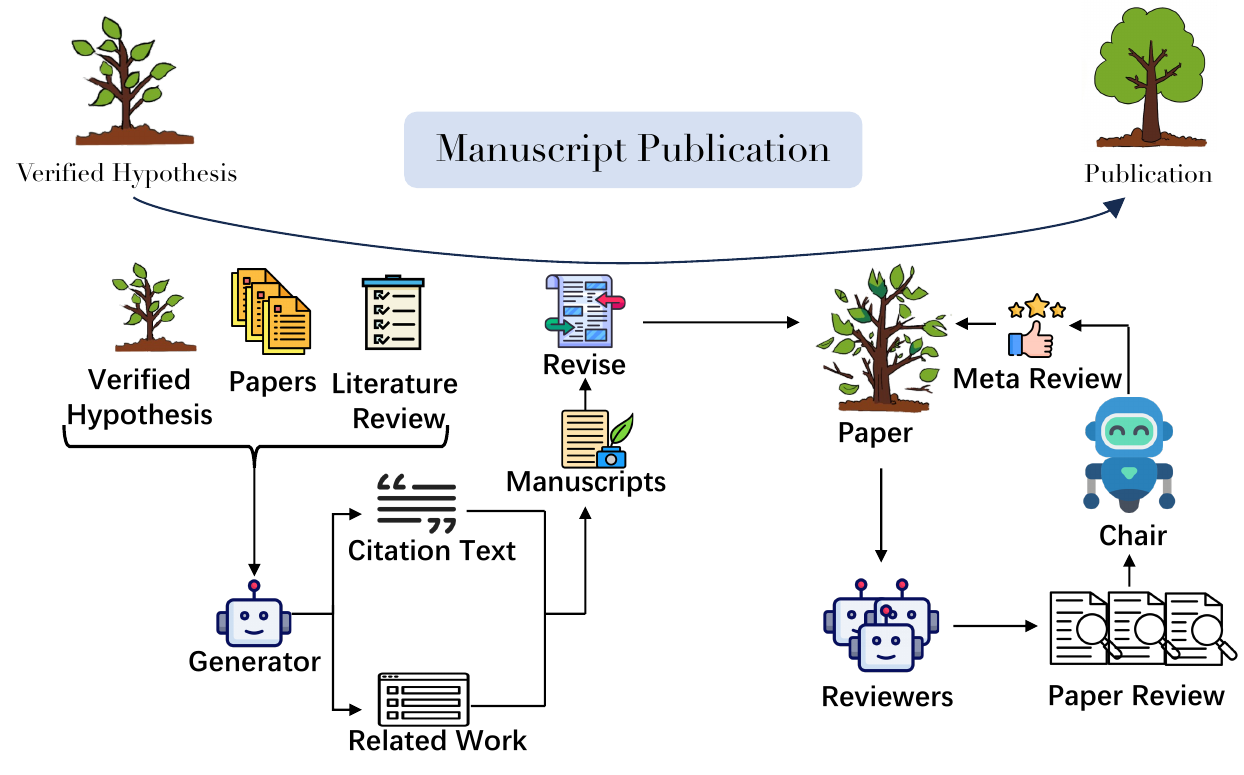}
        \caption{\label{figure:ManuscriptPublication}
        This figure shows the transformation of a validated hypothesis into a publication, leveraging outputs from the hypothesis formulation and validation stages.
        }

% \vspace{-5mm}
\end{figure}

Upon validating a hypothesis as feasible, researchers generally progress to the publication stage. As depicted in Figure~\ref{figure:ManuscriptPublication}, this section categorizes Manuscript Publication into two primary components: Manuscript Writing and Peer Review.

\subsection{Manuscript Writing} \label{sec:ManuscriptPublication_ManuscriptWriting}
Manuscript writing, also referred to as scientific or research writing. At this stage, researchers articulate the hypothesis they have formulated and the results they have validated in the form of a scholarly paper. This process is crucial, as it not only disseminates findings but also deepens researchers' understanding of their work ~\citep{09ColyarBecoming}. 

Early shared tasks focused on assisting researchers in writing or analyzing linguistic features~\citep{dale2010helping, daudaravicius2015automated}. Recent advances have led to three main directions: citation text generation, related work generation, and complete manuscript generation.

\paragraph{Citation Text Generation (Sentence Level)} A subset of research on AI in scientific writing has focused on citation text generation, which addresses the academic need for referencing prior work while mitigating model inaccuracies ~\citep{2305TianyuGaoEnabling, 2306NianlongGuSciLit}. For instance, ~\citet{2206YifanWangDisenCite} developed an automated citation generation system by integrating manuscript content with citation graphs. However, its reliance on rigid template-based architectures led to inflexible citation formats. This limitation motivated subsequent studies to propose incorporating citation intent as a control parameter during text generation, aiming to improve contextual relevance and rhetorical adaptability ~\citep{22YuMengxiaScientific, 22JungShingYunIntent, 2304RyanKooDecoding, 24GuNianlongControllable}.

\paragraph{Related Work Generation (Paragraph Level)} In contrast to citation text generation, several studies have focused on related work generation in scholarly writing, emphasizing the production of multiple citation texts and the systematic analysis of inter-citation relationships ~\citep{2201XiangciLiAutomatic,2404XiangciLiRelated}. The ScholaCite framework ~\citep{2402MartinBoyleAnnaShallow} leveraged GPT-4 to cluster citation sources and generate draft literature review sections, although it required manually provided reference lists. By contrast, the UR3WG system ~\citep{23ZhengliangShiTowards} adopted a retrieval-augmented architecture integrated with large language models to autonomously acquire relevant references. To improve the quality of generated related work sections,  ~\citet{24LuyaoYuReinforced} utilized GNNs to model complex relational dynamics between target manuscripts and cited literature, while ~\citet{24NishimuraToward} initiative advocated for explicit novelty assertions regarding referenced publications.  

\paragraph{Complete Manuscripts Generation (Full-text Level)} The aforementioned investigations primarily focused on specific components of scientific writing, while a study by ~\citet{2408YuxuanLaiInstruct} explored the progressive generation of complete manuscripts via structured workflows. The AI-Scientist system ~\citep{2408ChrisLuThe} further introduced section-wise self-reflection mechanisms to enhance compositional coherence. Several studies emphasized human-AI collaborative frameworks for improving writing efficiency ~\citep{2310LinZhichengTechniques, 2412KJKevinFengCocoa, 2404TalIfarganAutonomous}, whereas ~\citet{2406XiangruTangStep} concentrated on enabling personalized content generation in multi-author collaborative environments. Following initial manuscript drafting, subsequent text revision processes were systematically examined ~\citep{DuR0KLK22, 2303LaneJourdanText, DangSBG25}. The OREO system ~\citep{2204JingjingLiText} utilized attribute classification for iterative in-situ editing, while ~\citet{2204WanyuDuRead, 24MiltonPividoriA} incorporated researcher feedback loops for progressive text optimization. Notably, ~\citet{KimDR0K22, 2405EricChamounAutomated, 2306MikeDArcyARIES} proposed replacing manual feedback with automated evaluation metrics.

\subsection{Peer Review} \label{sec:ManuscriptPublication_PeerReview}
Peer review serves as a critical mechanism for improving the quality of academic manuscripts through feedback and evaluation, forming the cornerstone of quality control in scientific research. However, the process is hindered by its slow pace, high time consumption, and increasing strain due to the growing academic workload ~\citep{2111JialiangLinAutomated, 23KayvanKoushaArtificial, 2411MikeThelwallEvaluating}. To address these challenges and enhance manuscript quality, researchers have investigated the application of AI in peer review ~\citep{22WeizheYuanCan, 2306RyanLiuReviewerGPT, 2309LiangNiuUnveiling, 2405IliaKuznetsovWhat, thakkar2025can}. Peer review can be categorized into two main types: paper review generation and meta-review generation.

\paragraph{Paper Review Generation} In paper review generation, reviewers provide both scores and evaluations for manuscripts. For instance, ~\citet{22BasukiSetioThe} formulated score prediction as a regression task, ~\citet{22PanitanMuangkammuenExploiting} utilized semi-supervised learning, and ~\citet{2406PauloHenriqueCoutoRelevAI} treated the task as a classification problem to evaluate the alignment between manuscripts and review criteria. While these approaches focused on label prediction for paper reviews, ~\citet{22WeizheYuanKID} extended the scope by directly generating reviews through the construction of a concept graph integrated with a citation graph.

Subsequently, a pilot study conducted by ~\citet{2307ZacharyRobertsonGPT4} demonstrated the capability of GPT-4 to generate paper reviews. Further investigations, such as those by AI-Scientist ~\citep{2408ChrisLuThe} and ~\citet{2310WeixinLiangCan}, evaluated its performance as a review agent. Additionally, systems like MARG ~\citep{2401MikeDArcyMARG} and SWIF2T ~\citep{2405EricChamounAutomated} employed multi-agent frameworks to generate reviews via internal discussions and task decomposition. In contrast, AgentReview ~\citep{2406YiqiaoJinAgentReview} and ~\citet{2406ChengTanPeer} modeled the review process as a dynamic, multi-turn dialogue. Furthermore, CycleResearcher ~\citep{2411YixuanWengCycleResearcher} and OpenReviewer ~\citep{2412MaximilianIdahlOpenReviewer} fine-tuned models for comparative reviews and structured outputs aligned with conference guidelines.

\paragraph{Meta-Review Generation} In meta-review generation, chairs are tasked with identifying a paper's core contributions, strengths, and weaknesses while synthesizing expert opinions on manuscript quality. Meta-reviews are conceptualized as abstractions of comments, discussions, and paper abstracts ~\citep{2305MiaoLiSummarizing}. ~\citet{2402ShubhraKantiKarmakerSantuPrompting} investigated the use of LLMs for automated meta-review generation, while ~\citet{2305QiZengMeta} proposed a guided, iterative prompting approach. MetaWriter ~\citep{24LuSunMetaWriter} utilized LLMs to extract key reviewer arguments, whereas GLIMPSE ~\citep{2406MaximeDarrinGLIMPSE} and ~\citet{2310SandeepKumarWhen} focused on reconciling conflicting statements to ensure fairness. Additionally, ~\citet{2402MiaoLiA} introduced a three-layer sentiment consolidation framework for meta-review generation, and PeerArg ~\citep{2409PurinSukpanichnantPeerArg} integrated LLMs with knowledge representation to address subjectivity and bias via a multiparty argumentation framework (MPAF). DeepReview~\citep{zhu2025deepreview} generates a comprehensive meta-review by simulating expert evaluation across multiple dimensions.

During the Manuscript Publication phase, researchers can leverage AI to systematically complete manuscript writing by incorporating validated hypothesis, related papers, and literature reviews. The manuscript is subsequently subjected to peer review, involving iterative revisions before culminating in its final publication.

\section{Benchmarks} \label{sec:Benchmarks}
Given that AI for research spans multiple disciplines, the tasks addressed within each domain vary significantly. To facilitate cross-domain exploration, we provide a summary of benchmarks associated with various areas, including research paper recommendation, systematic literature review, hypothesis generation, scientific claim verification, theorem proving, experiment verification, manuscript writing, and peer review.  An overview of these benchmarks is presented in Table~\ref{tab:benchmarks}.

\begin{table*}[!t]
\centering
%\small
\tiny
\setlength{\tabcolsep}{3.5pt}
\resizebox{\textwidth}{!}{
\begin{tabular}{llccccc}
\toprule
 Task & Benchmark & Domain & Size & Input & Output & Metric \\
\midrule
 & SCHOLAT \cite{20WeishengLiAcademic} & Research Paper Recommendation & 34,518 & -  & - & Recall, Precission, F1-score \\  
 & ACL selection network \cite{tao2020paper} & Research Paper Recommendation  & 18,718 & Topics & Related Papers & Accuracy \\
 & CiteSeer \cite{21Kanghybrid} & Research Paper Recommendation & 1,100 & Paper & Related Papers & Correlation Coefficient\\
 & SciReviewGen \cite{23TetsuKasanishiSciReviewGen} & Systematic Literature Review & 10,000+ & Abstracts & literature review & ROUGE \\
 & FacetSum \cite{21RuiMengBringing} & Systematic Literature Review & 60,024 & Source Text+Facet &  Summary of Facet & ROUGE  \\
 & BigSurvey \cite{22ShuaiqiLiuGenerating} & Systematic Literature Review & 7,000+ & Abstracts & Survey Paragraph & ROUGE, F1-score\\
 & SCHOLARQABENCH \cite{2411AkariAsaiOpenScholar} & Systematic Literature Review & 2,200 & Question & Answer with Citations & \makecell{Accuracy, Coverage, Citations\\+ Relevance, Usefulness}  \\
 & HiCaD \cite{2304KunZhuHierarchical} & Systematic Literature Review & 7,600 &  Reference Papers & Catalogues & \makecell{Catalogue Edit Distance Similarity (CEDS)\\+ Catalogue Quality Estimate (CQE)}  \\
 & CLUSTREC-COVID \cite{2408UriKatzKnowledge} & Systematic Literature Review & 2,284 &  Titles, Abstracts & Topic &  Clusters per Topic \\
 & CHIME \cite{2407ChaoChunHsuCHIME}  & Systematic Literature Review & 2,174 & Topic & Hierarchies & F1-score \\
 & \citet{24DerekFWongOverview} & Systematic Literature Review & 700 & Subject, Reference & Title,Content & - \\ 
 & MASSW \cite{2406XingjianZhangMASSW}  & Hypothesis Generation &  152000 & Context of Literature & Hypothesis & \makecell{BLEU, ROUGE, BERTScore,\\+ Cosine Similarity, BLEURT} \\
 & IdeaBench \cite{2411SikunGuoIdeaBench}  & Hypothesis Generation & 2,374 &  Instruction, Background Information & Hypothesis & \makecell{Insight Score, BERTScore, Novelty, \\+ LLM Similarity Rating, Feasibility}\\
 & SCIMON \cite{2305QingyunWangSciMON} & Hypothesis Generation & - & Background Context  & Idea & \makecell{ROUGE, BERTScore\\
 +BARTScore, Novelty} \\
 & MOOSE\citet{2309ZonglinYangLarge}  & Hypothesis Generation & 50 & Background, Inspiration  & Hypothesis & \makecell{Validness, Novelty\\+ Helpfulness} \\
 & DISCOVERYBENCH \cite{2407DiscoveryBenchBodhisattwa}  & Hypothesis Generation & 1,167 & Data & Discovery & \makecell{Hypothesis Match Score} \\
\multirow{-23}{*}{Hypothesis} & LiveIdeaBench \cite{2412LiveIdeaBench}  & Hypothesis Generation & - & Scientific Keywords & Idea & \makecell{Originality, Feasibility\\+ Fluency, Flexibilit} \\
\multirow{-23}{*}{Formulation} & \citet{2409SandeepKumarCan}  & Hypothesis Generation & 100 & Paper without Future Work & Idea & \makecell{Idea Alignment Score, Idea Distinctness Index} \\

\midrule
 & SciRIFF \cite{2406DavidWaddenSciRIFF} &  Scientific Claim Verification & 137,000 & Evidence, Task prompt & Structured Paragraph &  F1, BLEU \\
 & SCIFACT \cite{2004DavidWaddenFact} & Scientific Claim Verification & 1,409 & Claim, Evidence & Rationale Sentences, Label & Precision, Recall, Micro-F1 \\
 & SCIFACT-OPEN \cite{WaddenLKCBWH22} & Scientific Claim Verification & 279 &  Claim, Evidence  & Rationale Sentences, Label & Precision, Recall,Micro-F1 \\
 & MISSCI \cite{Glockner0NG24} & Scientific Claim Verification & 435 & Claim, Premise, Context  & Verification & \makecell{Micro F1-score,P@1,Arg@1\\+ METEOR Score,BERTScore\\ +NLI-A, NLI-S, Matches@1} \\
 & FEVER \cite{ThorneVCM18} & Scientific Claim Verification & 185,445 & Claim, Evidence  & Label, Necessary Evidence & \makecell{F1-Score,Oracle Accuracy\\+ Accuracy,Recall} \\
 & XClaimCheck \cite{24WeiYuKaoMAGIC}  & Scientific Claim Verification & 16,177 & Claim, Evidence  & Label, Argument & Macro-F1, Accuracy \\
 & HEALTHVER \cite{SarroutiAMD21} & Scientific Claim Verification & 14330 & Claim, Evidence  & Label & \makecell{Macro Precision, Macro Recall\\ + Macro F1-score, Accuracy} \\
 & QuanTemp \cite{VAAS24} & Scientific Claim Verification & 15,514 & Claim, Evidence  & Label & \makecell{Weighted-F1 Score, Macro-F1, BLEU, \\+ BERTScore, Cohen’s Kappa Score\\+ Human Evaluation} \\
 & SCITAB \cite{LuPLNK23} & Scientific Claim Verification & 1,225 & Claim, Evidence  & Label & \makecell{Macro-F1} \\
 & Check-COVID \cite{WangHCASM23} & Scientific Claim Verification & 1,504 & Claim & Evidence & \makecell{Accuracy, Precision, Recall, Macro-F1} \\
 & HealthFC \cite{VladikaSM24} & Scientific Claim Verification & 750 & Claim, Evidence  & Label & \makecell{Precision, Recall, F1-Macro} \\
 & FACTKG \cite{2305JihoKimFactKG} & Scientific Claim Verification & 108,000 & Claim, Evidence  & Label & \makecell{ Accuracy} \\
 & BEAR-FACT \cite{2402AmelieWhrlWhat} & Scientific Claim Verification & 1,448 & \makecell{Claim, Evidence\\ +Entity/Relation Information}  & Label & \makecell{F1-Score} \\
 & MINIF2F \cite{ZhengHP22} & Theorem Proving  & 488 & Problem, Theorem & Proof & \makecell{Pass Rate } \\
 & FIMO \cite{2309ChengwuLiuFIMO} & Theorem Proving  & 149 & Problem, Theorem, statements & Proof & \makecell{Pass Rate } \\
 & LeanDojo \cite{YangSGCSYGPA23} & Theorem Proving  & 98,734 & Problem, Theorem & Proof & \makecell{ R@k, MRR, Pass Rate} \\
 & Lean-github \cite{2407LEANGitHub} & Theorem Proving  & 28,597 & Problem, Theorem & Proof & \makecell{Accuracy, Pass Rate} \\
 & TRIGO-real \cite{XiongSYWYLLGCHZ23} & Theorem Proving  & 427 & Problem, Theorem & Proof & \makecell{ Pass Rate, Accuracy, EM@n} \\
 & TRIGO-web \cite{XiongSYWYLLGCHZ23} & Theorem Proving  & 453 & Problem, Theorem & Proof & \makecell{Pass Rate, Accuracy, EM@n } \\
 & TRIGO-gen \cite{XiongSYWYLLGCHZ23} & Theorem Proving  & - & Problem, Theorem & Proof & \makecell{Pass Rate, Accuracy, EM@n  } \\
 & CoqGym \cite{YangD19} & Theorem Proving  & 71,000 & Problem, Theorem & Proof & \makecell{Success Rate} \\
 & MLAgentBench \cite{2310QianHuangMLAgentBench} & Experiment Validation  & 13 & - & - & \makecell{Competence, Efficiency} \\
 & AAAR-1.0 \cite{2410RenzeLouAAAR10} & Experiment Validation  & - & Instance, Papers & Design, Explanation & \makecell{ S-F1, S-Precision, S-Recall\\+ S-Match, ROUGE} \\
 & TASKBENCH \cite{0001STZRY00Z24} & Experiment Validation  & 17,331 & - & - & \makecell{ROUGE,  t-F1, v-F1\\ +Normalized Edit Distance } \\
 & Spider2-V \cite{CaoLWCFGXZHMXXZ24} & Experiment Validation  & 494 & Task & Experiment Execution & \makecell{Success Rate } \\
 & CORE-Bench \cite{2409ZacharySSiegelCOREBench} & Experiment Validation  & 270 & Task Requirements & Experiment Result & \makecell{Accuracy} \\
 & LAB-Bench \cite{2407LABBench} & Experiment Validation  & 2400 & Multiple-choice Question & Answer & \makecell{Accuracy, Precision, Coverage}\\ 
 & PaperBench \cite{starace2025paperbench} & Experiment Validation  & 20 & Paper, Additional Information & Code & \makecell{Replication Score}\\
\multirow{-38}{*}{Hypothesis} & SUPER \cite{BoginYG0BCSK24} & Experiment Validation  & 801 & Task Requirements & - & \makecell{Accuracy, Landmark-Based Evaluation} \\
\multirow{-38}{*}{Validation} & ScienceAgentBench \cite{ChenCNZWYLLWLDX25} & Experiment Validation  & 102 & \makecell{Task Instruction, Dataset Information\\ +Expert-Provided Knowledge} & Program & \makecell{Valid Execution Rate, Success Rate, CodeBERTScore, API Cost} \\

\midrule
 & SciCap+ \cite{23ZhishenYangSciCap} & Manuscript Writing & 414,000 & \makecell{Figure, OCR tokens \\+ Mention Paragraph} & Caption & \makecell{BLEU, ROUGE, METEOR\\+ CIDEr, SPICE} \\
 & AAN Corpus \cite{RadevMQA13} & Manuscript Writing & - & - & - & - \\
 & SciSummNet \cite{YasunagaKZFLFR19} & Manuscript Writing & 1,000 & Paper,Citation Sentence & Summary & \makecell{ROUGE} \\
 & CiteBench \cite{FunkquistK0G23} & Manuscript Writing & 358,765 & Cited Papers, Context & Citation Text & \makecell{ROUGE, BERTScore} \\
 & ALCE \cite{2305TianyuGaoEnabling} & Manuscript Writing & 3,000 &  Question & Answer with Citations & \makecell{Recall, Precision} \\
 & GCite \cite{2206YifanWangDisenCite} & Manuscript Writing & 2,500 & Citing/Cited Paper & Citation Text & \makecell{BLEU, ROUGE} \\
 & ARXIVEDITS \cite{JiangXS22} & Manuscript Writing & 1,000 & Sentence Pairs & Sentence, Intent & \makecell{Precision,Recall,F1-score}\\
 & CASIMIR \cite{JourdanBHD24} & Manuscript Writing & 15,646 & Original Sentence & Revised Sentence &  \makecell{Exact-match (EM),SARI, BLEU,\\+ ROUGE-L,Bertscore}\\
 & ParaRev \cite{jourdan2025pararev} & Manuscript Writing & 48,203 & Original Paragraph & Revised Paragraph &  \makecell{ROUGE-L,SARI\\+ BertScore}\\
 & SCHOLAWRITE \cite{2502LingheWangScholaWrite} & Manuscript Writing & 62,000 & Before-text & Writing Intention, After-text &  \makecell{F1-score, Lexical Diversity, Topic Consistency, Intention Coverage}\\
 & MReD \cite{ShenCZBYS22} & Peer Review & 7,089 & Reviews & Meta-Review & \makecell{ROUGE} \\
 & ORSUM\cite{zeng2024scientific} & Peer Review & 15,062 &Reviews &Meta-Review & \makecell{ROUGE-L, BERTScore, FACTCC\\+ SummaC, DiscoScore} \\ 
 & PeerRead v1 \cite{KangADZKHS18} & Peer Review & 107,000 & Reviews & Accept/Reject & \makecell{ Accuracy} \\
 &NLPeer  \cite{DyckeKG23} & Peer Review & 5,000 & Reviews,Paper & \makecell{Review Score, Connection, \\+ Review Category} & \makecell{MRSE, F1-macro\\+ Precision, Recall} \\
 &AMPERE \cite{HuaNBW19} & Peer Review & 400 & Review & Review with Type & \makecell{ Precision, Recall, F1-score} \\
 &MOPRD  \cite{LinSZCS23} & Peer Review & 6,578 & Reviews,Paper & \makecell{Editorial Decision, Review, \\+ Meta-Review, Author Rebuttal} & \makecell{ROUGE, BARTScore} \\
 &ARIES  \cite{2306MikeDArcyARIES} & Peer Review & 1,720 & Review Comment, Edits &  Comment-Edit Pairs & \makecell{Precision, Recall, F1-score} \\
 &ASAP-Review \cite{22WeizheYuanCan} & Peer Review & - & Paper & Review & \makecell{Aspect Coverage, Aspect Recall, \\+Semantic Equivalence
\\+Human: Recommendation Accuracy(RAcc),\\+Informativeness(Info),Aspect-level,\\+Constructiveness(ACon) and Summary accuracy} \\
 & ReviewMT  \cite{2406ChengTanPeer} & Peer Review & 26,841 & Paper & Review Dialogue  & \makecell{ROUGE,BLEU,METEOR} \\
\multirow{-30}{*}{Manuscript} &ReAct  \cite{ChoudharyMM21} & Peer Review & 6,250 & Review  & Classification of Review  & \makecell{Accuracy} \\
\multirow{-30}{*}{Publication}  &PEERSUM \cite{2305MiaoLiSummarizing} & Peer Review & - &Reviews  &Meta-Review  & \makecell{ROUGE,BERTScore,UniEval,ACC} \\

\bottomrule
\end{tabular}
} % resizebox
\caption{An overview of benchmarks on AI for research. In the Input, Output, and Metric columns, the '+' symbol indicates that the row is a continuation of the previous row.}
\label{tab:benchmarks}
\end{table*}

\section{Tools} \label{sec:Tools}
To accelerate the research workflow, we have curated a collection of tools designed to support various stages of the research process, with their applicability specified for each stage. To ensure practical relevance, our selection criteria emphasize tools that are publicly accessible or demonstrate significant influence on GitHub. A comprehensive overview of these tools is presented in Table~\ref{tab:tools}.

\section{Future Directions} \label{sec:Challenges}

We identify several intriguing and promising avenues for future research.
\subsection{Integration of Diverse Research Tasks} \label{sec:Challenges_Diverse}
The research process is an integrated pipeline of interdependent stages. Paper recommendation and literature review provide an AI tool with a field overview and relevant works, ensuring that hypothesis generation is informed and of higher quality. Hypothesis validation assesses feasibility both logically and practically, with results feeding back to refine the hypothesis~\citep{penades2025ai}. In manuscript writing, validated hypotheses and prior outputs serve as key inputs. Peer review evaluates the manuscript and offers feedback across modules, enabling the hypothesis generator to adjust content accordingly~\citep{2408ChrisLuThe}. In addition, combinations can also be made between some small fields, for instance, meta-review generation could be integrated with scientific claim verification, experiment verification could be linked with hypothesis formulation~\citep{jansen2025codescientist, 2501YuanJiakangDolphin, 2411ZijunLiuAIGS}, and research paper recommendation systems could be connected with manuscript writing processes~\citep{2306NianlongGuSciLit}. Furthermore, some studies have begun to emphasize the development of systems capable of covering multiple stages of the research process~\citep{2406PeterAJansenDiscoveryWorld,2411YixuanWengCycleResearcher,2412HaofeiYuResearchTown}.

\subsection{Integration with Reasoning-Oriented Language Models} \label{sec:Challenges_ReasoningOriented}
Research is a process that places a significant emphasis on logic and reasoning. Theorem proving serves as a subtask within logical reasoning~\citep{2404ZhaoyuLiA}, while hypothesis generation is widely recognized as the primary form of reasoning employed by scientists when observing the world and proposing hypothesis to explain these observations~\citep{2309ZonglinYangLarge}. Experiment verification, in turn, demands a high degree of planning capability from models~\citep{2402SubbaraoKambhampatiPosition}. Recent advances in reasoning-oriented language models, such as OpenAI-o1~\citep{2412AaronJaechOpenAIo1} and DeepSeek-R1~\citep{2501deepseekAIDeepseek-r1}, have substantially enhanced the reasoning abilities of these models. Consequently, we posit that integrating reasoning language models with reasoning tasks is a promising future direction. This prediction was validated by experiments conducted by ~\citet{2501SchmidgallAgent} using o1-Preview.

Furthermore, in Appendix ~\S\ref{sec:appendixchallenges}, we provide a summary of the challenges in hypothesis formulation, validation, and manuscript publication.

\section{Ethical Considerations}
\label{sec:Ethical Considerations}
AI has demonstrated significant potential in enhancing productivity by mitigating human limitations, thereby motivating increased investigation into its capacity to accelerate the research process ~\citep{2024MesseriArtificial}. Nevertheless, its integration into scientific workflows introduces a range of ethical concerns ~\citep{2306BenediktFecherFriend, 2304MeredithScientists}, including algorithmic biases, data privacy issues, risks of plagiarism, and the broader implications of AI-generated content for research communities. In this work, we examine these ethical challenges across the key stages of the research lifecycle: hypothesis formulation, validation, and publication.

During the hypothesis formulation stage, research paper recommendation systems and literature reviews are commonly employed; however, they often suffer from limitations that can lead to the formation of information bubbles and restrict exposure to diverse viewpoints. Furthermore, these systems tend to reinforce recognition disparities between prominent and lesser-known researchers and may inadvertently contribute to the dissemination of misinformation ~\citep{Polonioli21, 2402FranciscoBolaArtificial}. To address these biases, recommendation algorithms can be enhanced by emphasizing content-based rather than author-based recommendations and by incorporating robust evaluation mechanisms to strengthen the credibility of suggested materials.

In contrast, AI-driven hypothesis generation presents more pronounced ethical challenges. First, the attribution of intellectual property rights and authorship for AI-generated hypotheses remains ambiguous ~\citep{2402BodhisattwaPrasadMajumder}. Additionally, the widespread generation of low-quality content poses a risk of diluting the integrity of the academic landscape~\citep{2410XiangHuNova}, while the potential misuse of such technologies for illicit purposes cannot be overlooked ~\citep{2409ChengleiSiCan}. Addressing these concerns necessitates the development of robust accountability frameworks, the assignment of clear responsibility for AI-generated outputs to researchers, and the establishment of appropriate legal and regulatory mechanisms.

During the hypothesis validation phase, automated systems for scientific fact-checking remain underdeveloped. This limitation may be exploited by malicious actors to create advanced misinformation generators capable of circumventing existing fact-checking tools ~\citep{2112DavidWaddenMultiVerS}. Likewise, in the context of experimental validation, there is a risk of unethical or legally questionable experiments being designed ~\citep{2502TransformingSteffen}. These concerns underscore the need for continued research into model safety.

During the manuscript publication stage, several challenges remain. Text generated by AI models may carry a risk of plagiarism ~\citep{23SalvagnoCan, 2502TarunGuptaAll}, while AI-assisted peer reviews often offer vague feedback and exhibit inherent biases ~\citep{2309LaurieA, 24IddoDroriHuman, 2502PataranutapornCan}. To address these issues, the development of robust detection methods is essential. However, current detection tools are still in the early stages of maturity ~\citep{2502TarunGuptaAll}.

\section{Conclusion} \label{sec:Conclusion}

This paper provides a systematic survey of existing research on AI for research, offering a comprehensive review of the advancements in the field. Within each category, we offer detailed descriptions of the associated subfields. In addition, we examine current challenges, ethical considerations, and potential directions for future research. To support researchers in exploring AI-driven research applications and enhancing workflow efficiency, we also summarize existing benchmarks and tools, accompanied by a comparative analysis of representative methods and their capabilities.

Furthermore, in the course of investigating various subfields within AI for research, we observed that this domain remains in its infancy. Research in numerous directions remains at an experimental stage, and substantial progress is necessary before these approaches can be effectively applied in practical scenarios. We hope that this survey serves as an introduction to the field for researchers and contributes to its continued advancement.

\section*{Limitation}
This study presents a comprehensive survey of AI for research, based on the framework of the research process conducted by human researchers.

We have made our best effort, but there may still be some limitations. Due to space constraints, we provide only concise summaries of each method without detailed technical elaboration. Given the rapid progress in AI and the expanding research landscape, we primarily focus on works published after 2022, with earlier studies receiving less attention. To emphasize areas that closely mimic the human research process, some topics are excluded from the main text but briefly discussed in Appendix~\S\ref{sec:discussion}. Moreover, as AI for Research is still an emerging field, the lack of standardized benchmarks and evaluation metrics hinders direct comparison. Nonetheless, we offer a comparative analysis of representative methods across domains using attribute graphs in Appendix~\S\ref{sec:Ability Comparison}.

\section*{Acknowledgments}
Xiachong Feng and Xiaocheng Feng are the co-corresponding authors of this work. We thank the anonymous reviewers for their insightful comments. This work was supported by the National Natural Science Foundation of China (NSFC) (grant 62276078, U22B2059), the Key R\&D Program of Heilongjiang via grant 2022ZX01A32, and the Fundamental Research Funds for the Central Universities ( XNJKKGYDJ2024013 ).

\onecolumn
\renewcommand{\arraystretch}{1.5} % 减小行距

\begin{table*}
\tiny % 保持最小字号
\resizebox{\textwidth}{!}{
\begin{tabular}{p{2cm}p{1.65cm}p{1cm}p{1.2cm}p{1.4cm}p{1cm}p{1.3cm}p{1cm}p{0.7cm}p{1cm}}
\hline
\textbf{Tool} & \textbf{Research Paper Recommendation} & \textbf{Systematic Literature Review} & \textbf{Hypothesis Generation} & \textbf{Scientific Claim Verification} & \textbf{Theorem Proving} & \textbf{Experiment Verification} & \textbf{Manuscript Writing} & \textbf{Peer Review} & \textbf{Reading Assistance} \\

\hline
\href{https://www.connectedpapers.com/}{Connected Paper} &\highlightcheck & & & & & & & &\\
\hline
\href{https://inciteful.xyz/}{Inciteful} &\highlightcheck & & & & & & & &\\
\hline
\href{https://www.litmaps.com/}{Litmaps} &\highlightcheck & & & & & & & &\\
\hline
\href{https://pasa-agent.ai/}{Pasa} &\highlightcheck & & & & & & & &\\
\hline
\href{https://researchrabbitapp.com/}{Research Rabbit} &\highlightcheck & & & & & & & &\\
\hline
\href{https://www.semanticscholar.org/}{Semantic Scholar} &\highlightcheck & & & & & & & &\highlightcheck\\
\hline
\href{https://gengo.sotaro.io/}{GenGO} &\highlightcheck & & & & & & & &\highlightcheck\\
\hline
\href{https://jenni.ai/}{Jenni AI} &\highlightcheck & & & & & &\highlightcheck & &\highlightcheck\\
\hline
\href{https://elicit.com/}{Elicit} &\highlightcheck &\highlightcheck & & & & & & &\\
\hline
\href{https://undermind.ai/home/}{Undermind} &\highlightcheck &\highlightcheck & & & & & & &\\
\hline
\href{https://openscholar.allen.ai/}{OpenScholar} &\highlightcheck &\highlightcheck & & & & & & &\\
\hline
\href{https://researchbuddy.app/}{ResearchBuddies} &\highlightcheck &\highlightcheck & & & & & & &\\
\hline
\href{https://www.hyperwriteai.com/}{Hyperwrite} &\highlightcheck &\highlightcheck & & & & &\highlightcheck & &\\
\hline
\href{https://consensus.app/search/}{Concensus} &\highlightcheck &\highlightcheck & &\highlightcheck & & & & &\\
\hline
\href{https://iris.ai/}{Iris.ai} &\highlightcheck &\highlightcheck & &\highlightcheck & & & & &\highlightcheck\\
\hline
\href{https://mirrorthink.ai/}{MirrorThink} &\highlightcheck &\highlightcheck & &\highlightcheck & &\highlightcheck & & &\highlightcheck\\
\hline
\href{https://typeset.io/}{SciSpace} &\highlightcheck &\highlightcheck & & & & &\highlightcheck &\highlightcheck &\highlightcheck\\
\hline
\href{https://askyourpdf.com}{AskYourPDF} &\highlightcheck &\highlightcheck & &\highlightcheck & & &\highlightcheck &\highlightcheck &\highlightcheck\\
\hline
\href{https://sciai.las.ac.cn/}{Iflytek} &\highlightcheck &\highlightcheck & &\highlightcheck &\highlightcheck &\highlightcheck &\highlightcheck & &\highlightcheck\\
\hline
\href{https://platform.futurehouse.org/}{FutureHouse} &\highlightcheck &\highlightcheck & \highlightcheck & & &\highlightcheck & & &\\
\hline
\href{https://www.read.enago.com/}{Enago Read} &\highlightcheck &\highlightcheck &\highlightcheck &\highlightcheck &\highlightcheck & & & &\highlightcheck\\
\hline
\href{https://www.aminer.cn/}{Aminer} &\highlightcheck &\highlightcheck &\highlightcheck &\highlightcheck &\highlightcheck &\highlightcheck &\highlightcheck & &\highlightcheck\\
\hline
\href{https://github.com/GAIR-NLP/OpenResearcher}{OpenRsearcher} &\highlightcheck &\highlightcheck &\highlightcheck & &\highlightcheck &\highlightcheck &\highlightcheck &\highlightcheck &\highlightcheck\\
\hline
\href{https://rflow.ai/}{ResearchFlow} &\highlightcheck &\highlightcheck & &\highlightcheck &\highlightcheck &\highlightcheck &\highlightcheck &\highlightcheck &\highlightcheck\\
\hline
\href{https://you.com/?chatMode=research}{You.com} &\highlightcheck &\highlightcheck &\highlightcheck &\highlightcheck &\highlightcheck &\highlightcheck &\highlightcheck &\highlightcheck &\highlightcheck\\
\hline
\href{https://gptr.dev/}{GPT Researcher} & &\highlightcheck & & & & & & &\\
\hline
\href{https://picoportal.org/}{PICO Portal} & &\highlightcheck & & & & & & &\\
\hline
\href{https://surveyx.cn/list}{SurveyX} & &\highlightcheck & & & & & & &\\
\hline
\href{https://pharma.ai/dora}{Scinence42:Dora} & &\highlightcheck & & & & &\highlightcheck & &\\
\hline
\href{https://storm.genie.stanford.edu/}{STORM} & &\highlightcheck & & & & &\highlightcheck & &\\
\hline
\href{https://chatdoc.com/}{ChatDOC} & &\highlightcheck & & & & & & &\highlightcheck\\
\hline
\href{https://scite.ai/home}{Scite} & &\highlightcheck & & & & & & &\highlightcheck\\
\hline
\href{https://silatus.com/}{Silatus} & &\highlightcheck & & & & & & &\highlightcheck\\   
\hline
\href{https://github.com/SamuelSchmidgall/AgentLaboratory}{Agent Laboratory} & &\highlightcheck & & & &\highlightcheck &\highlightcheck & &\\
\hline
\href{https://sider.ai/}{Sider} & &\highlightcheck & & & & &\highlightcheck & &\highlightcheck\\
\hline
\href{https://quillbot.com/}{Quillbot} & &\highlightcheck & & & & &\highlightcheck &\highlightcheck &\highlightcheck\\
\hline
\href{https://scholarai.io/}{Scholar AI} & &\highlightcheck & &\highlightcheck & &\highlightcheck &\highlightcheck &\highlightcheck &\highlightcheck\\
\hline
\href{https://sakana.ai/ai-scientist/}{AI-Researcher} &  &\highlightcheck &\highlightcheck & & &\highlightcheck &\highlightcheck &\highlightcheck &\\
\hline
\href{https://sakana.ai/ai-scientist/}{AI Scientist} & & &\highlightcheck & & &\highlightcheck &\highlightcheck &\highlightcheck &\\
\hline
\href{https://isabelle.in.tum.de/}{Isabelle} & & & & &\highlightcheck & & & &\\
\hline
\href{https://github.com/lean-dojo/LeanCopilot}{LeanCopilot} & & & & &\highlightcheck & & & &\\
\hline
\href{https://github.com/wellecks/llmstep}{Llmstep} & & & & &\highlightcheck & & & &\\
\hline
\href{https://proverbot9001.ucsd.edu/}{Proverbot9001} & & & & &\highlightcheck & & & &\\
\hline
\href{https://github.com/ifyz/chatgpt_academic}{chatgpt\textunderscore academic} & & & & & &  &\highlightcheck & &\\ 
\hline
\href{https://github.com/binary-husky/gpt_academic}{gpt\_academic} & & & & & & &\highlightcheck & &\\
\hline
\href{https://coschedule.com/headline-analyzer}{HeadlineAnalyzer} & & & & & & &\highlightcheck & &\\
\hline
\href{https://editor.langsmith.co.jp/}{Langsmith Editor} & & & & & & &\highlightcheck & &\\
\hline
\href{https://textero.ai/}{Textero.ai} & & & & & & &\highlightcheck & &\\
\hline
\href{https://wordvice.ai/}{Wordvice.AI} & & & & & & &\highlightcheck & &\\
\hline
\href{https://writesonic.com/}{Writesonic} & & & & & & &\highlightcheck & &\\
\hline
\href{https://x.writefull.com/}{Writefull} & & & & & & &\highlightcheck &\highlightcheck &\\
\hline
\href{https://www.covidence.org/}{Covidence} & & & & & & & &\highlightcheck &\\
\hline
\href{https://www.penelope.ai/}{Penelope.ai} & & & & & & & &\highlightcheck &\\
\hline
\href{https://byte-science.com/}{Byte-science} & & & & & & & & &\highlightcheck\\
\hline
\href{https://papers.cool/}{Cool Papers} & & & & & & & & &\highlightcheck\\
\hline
\href{https://www.explainpaper.com/}{Explainpaper} & & & & & & & & &\highlightcheck\\
\hline
\href{https://uni-finder.dp.tech/question}{Uni-finder} & & & & & & & & &\highlightcheck\\
\hline

\end{tabular}
}
\caption{Tools for Research Paper Assistance}\label{tab:tools}
\end{table*}

\twocolumn

\bibliography{custom}

\appendix

\section{Further Discussion}
\label{sec:discussion}
\paragraph{Open Question: What is the difference between AI for science and AI for research?}
We posit that AI for research constitutes a subset of AI for science. While AI for research primarily focuses on supporting or automating the research process, it is not domain-specific and places greater emphasis on methodological advancements. In contrast, AI for science extends beyond the research process to include result-oriented discovery processes within specific domains, such as materials design, drug discovery, biology, and the solution of partial differential equations~\citep{2310YizhenZhengLarge,2311AI4ScienceThe,2406YuZhangA}.

\paragraph{Open Question: What is the difference between hypothesis generation and scientific discovery?}
Hypothesis generation, which is primarily based on literature-based review (LBD)~\citep{86DonRSwansonUndiscovered,17SebastianEmerging}, emphasizing the process by which researchers generate new concepts, solutions, or approaches through existing research and their own reasoning. Scientific discovery encompasses not only hypothesis generation, but also innovation in fields like molecular optimization and drug development ~\citep{2401GeyanYeDrugAssist,24ShengchaoLiuConversational}, driven by outcome-oriented results.

\paragraph{Open Question: What is the difference between systematic literature review and related work generation?}
Existing research frequently addresses the systematic literature survey, which constitutes a component of the knowledge synthesis process during hypothesis formulation, alongside the related work generation phase in manuscript writing~\citep{25LuoZimingLLM4SR}. However, we argue that these two tasks are distinct in nature. The systematic literature survey primarily focuses on summarizing knowledge extracted from diverse scientific documents, thereby assisting researchers in acquiring an initial understanding of a specific field~\citep{22NoufIbrahimAltmamiAutomatic}. In contrast, related work generation focuses on the writing process, emphasizing selection of pertinent literature and effective content structuring ~\citep{24NishimuraToward}.

\paragraph{Discussion: Potential links between artificial intelligence systems and human research practice}
\begin{itemize}[leftmargin=2em]
    \item In research paper recommendation, PaperWeaver~\citep{2403YoonjooLeePaperWeaver} offers an interactive page that allows users to modify the topics they are interested in. 
    \item In systematic literature review, ~\citet{24BlockWhat} highlights the significant role of humans, including setting correct questions and individualized problem-solving and theorizing. Meanwhile, ~\citet{2407ChaoChunHsuCHIME} emphasizes manual correction during the outline generation process.
    \item In hypothesis generation, AI engages more closely with human researchers, ranging from scenarios where humans provide the core ideas and AI contributes by iteratively refining them ~\citep{2410KevinPuIdeaSynth}, to more collaborative settings where humans and AI engage in dialogue to facilitate new scientific discoveries ~\citep{2411ZiqiNiMatPilot, 24ShengchaoLiuConversational, 2401GeyanYeDrugAssist}.
    \item In scientific claim verification, ~\citet{2305EnesAltuncuaedFaCT, 2301AnubrataDasThe} highlight the critical role of experts in countering fake scientific news and advocate for the incorporation of expert opinions as a form of evidence.
    \item In theorem proving, ~\citet{2404PeiyangSongTowards} proposes leveraging LLMs as assistants to human researchers by generating suggested proof steps throughout the proving process.
    \item In experiment Validation, ~\citet{2411ZiqiNiMatPilot} enhances the experimental setup through human-AI dialogue, whereas ~\citet{2408RuochenLiMLRCopilot} incorporates human input and real-time adjustments during the execution phase to optimize experimental design.
    \item In manuscript writing, ~\citet{2404TalIfarganAutonomous, 2412KJKevinFengCocoa, 2204WanyuDuRead} require human intervention to suggest improvements to AI-generated paragraphs and enhance their quality through interactive methods.
    \item In peer review, ~\citet{2310SandeepKumarWhen, 2406MaximeDarrinGLIMPSE} advocate for assigning the responsibility of generating meta-reviews to human researchers. The role of AI is to assist by identifying conflicts among reviewers’ opinions and supporting the chair in the scoring process, rather than independently assigning scores.

\end{itemize}
At present, AI for research remains in its early stages, and most systems still rely heavily on human priors and conventional research workflows. This is due to both the limitations of current models and the lack of trust from the research community. However, as model capabilities significantly improve, we may witness a paradigm shift. Systems like AlphaFold have already demonstrated that impactful scientific contributions can be made without fully replicating human research processes. In the future, AI may become an autonomous scientific agent, pursuing its own pathways of discovery, potentially letting humans learn from the model.

\paragraph{Discussion: The involvement of AI in manuscript writing}
The application of AI in manuscript writing has been accompanied by significant controversy. As LLMs demonstrated advanced capabilities, an increasing number of researchers began adopting these systems for scholarly composition ~\citep{2404WeixinLiangMapping, 23CatherineAGaoComparing}. This trend raised concerns within the academic community ~\citep{23SalvagnoCan}, with scholars explicitly opposing the attribution of authorship to AI systems ~\citep{23leeCan}. Despite these reservations, the substantial time efficiencies offered by this technology led researchers to gradually accept AI-assisted writing practices ~\citep{24GrudaThree, 23HuangJingshanThe, 23ChenChatGPT}.This shift ultimately led to formal guidelines issued by leading academic journals ~\citep{24GanjaviPublishers, 2502ZiyangXuPatterns}.

\paragraph{Discussion: Some areas that have not been discussed}
In addition to the eight areas discussed above, there are other lines of work that also aim to support scientific research, such as reading assistance~\citep{KangHSLWH20, HeadLKFSWH21, 2303KyleLoThe}, which helps researchers read academic papers; literature processing\footnote{\url{https://sdproc.org}}, which handles documents in various formats to provide effective data for subsequent tasks; as well as code and data generation~\citep{bauer2024comprehensive, zheng2023codegeex}, which serve as a foundation for experimental validation. However, as our focus is on the core process of scientific research, we have chosen to omit these aspects from the main text.

\paragraph{ Discussion: Unified system and domain-specific system to automate research}

There is a clear distinction between unified and domain-specific AI research systems. Some efforts aim to develop general-purpose frameworks capable of supporting scientific discovery across domains (e.g., AI Scientist~\citep{2408ChrisLuThe}), while others target domain-specific challenges (e.g., AI in biology~\citep{irons2024ai}). Given the current limitations of AI capabilities, general-purpose systems have not yet fully replaced domain-specific approaches. Both directions remain valuable, but a long-term vision may favor general systems, as they hold the potential to integrate cross-disciplinary knowledge and push the boundaries of scientific understanding.

\section{Challenges}
\label{sec:appendixchallenges}
\subsection{Hypothesis Formulation} \label{sec:appendixchallenges_HF}
\paragraph{Knowledge Synthesize} Existing paper recommendation tools predominantly rely on the metadata of existing publications to suggest related articles, which often results in a lack of user-specific targeting and insufficiently detailed presentation that hampers comprehension. Leveraging LLMs can facilitate the construction of dynamic user profiles, enabling personalized literature recommendations and enhancing the richness of the contextual information provided for each recommended article, ultimately improving the user experience. In the process of generating systematic literature reviews, our practical experience reveals that the outline generation tools often produces redundant results with insufficient hierarchical structure. Moreover, the full-text generation process is prone to hallucinations—for instance, statements may not correspond to the cited articles—a pervasive issue in large language models~\citep{2311LeiHuangA,2402FranciscoBolaArtificial,2404Teosusnjakautomating}. This problem can be ameliorated by enhancing the foundational model capabilities or by incorporating citation tracing.

\paragraph{Hypothesis Generation} Most existing tools generate hypotheses by designing prompts or constructing systematic frameworks, which heavily rely on the capabilities of pre-trained models. However, these methods struggle to balance the novelty, feasibility, and validity of the hypotheses~\citep{2412RuochenLiLearning}.Furthermore, our investigation reveals that many current approaches adopt novelty and feasibility as evaluation metrics; these metrics are either difficult to quantify or require manual scoring, which can introduce bias. To date, there is no unified benchmark to compare the various methods, and we believe that future research should prioritize establishing a unified metric that objectively reflects the strengths and weaknesses of different approaches.

\subsection{Hypothesis Validation} \label{sec:appendixchallenges_HV}
Most existing scientific claim verification tools are largely confined to specific domains, exhibiting poor generalizability, which limits their practical applicability ~\citep{2305JurajVladikaScientific}. Theorem proving, the scarcity of relevant data adversely affects performance improvements through training , results across different proof assistants are not directly comparable, and the lack of standardized evaluation benchmarks presents numerous challenges. Moreover, current approaches remain predominantly in the research stage and lack practical tools that facilitate interaction with researchers ~\citep{2404ZhaoyuLiA}. Experiment Validation, as automatically generated experiments often suffer from a lack of methodological rigor, practical feasibility, and alignment with the original research objectives ~\citep{2410RenzeLouAAAR10}. All these fields require rigorous logical reasoning, and I believe that the recent surge in advanced reasoning technologies could potentially address these issues.

\subsection{Manuscript Publication} \label{sec:appendixchallenges_MP}
Similar to systematic literature surveys, manuscript writing is also adversely affected by hallucination issues~\citep{23AthaluriExploring,2311LeiHuangA}. Even when forced citation generation is employed, incorrect references may still be introduced~\citep{2024AljamaanReference}. Furthermore, the text generated by models requires meticulous examination by researchers to avoid ethical concerns, such as plagiarism risks~\citep{23SalvagnoCan}. AI-generated manuscript reviews frequently provide vague suggestions and are susceptible to biases~\citep{2405EricChamounAutomated,24IddoDroriHuman}. Additionally, during meta-review generation, models can be misled by erroneous information arising from the manuscript review process~\citep{2310SandeepKumarWhen}. To address these issues, it may be necessary for the industry to establish appropriate regulations or to employ AI-based methods for detecting AI-generated papers and reviews~\citep{2111JialiangLinAutomated}.

\section{Ability Comparison}
\label{sec:Ability Comparison}
An effective survey should not only summarize existing methods within a field but also provide comparative analyses of different approaches. However, the domain of AI for Research remains in its early stages, with many areas lacking standardized benchmarks and even established evaluation metrics. To facilitate a clearer understanding of the distinctions among various methods, we draw on existing literature~\citep{2302HyeonsuB.KangComLittee, 2402FranciscoBolaArtificial, 25LuoZimingLLM4SR, 2305JurajVladikaScientific, 2303ZonglinYangLogical, 2201XiangciLiAutomatic, 2404XiangciLiRelated, 2111JialiangLinAutomated} and adopt attribute graphs to compare representative approaches within each subfield, as illustrated in table ~\S\ref{tab:research_paper_recommendation} to table ~\S\ref{tab:peer_review}. 

\begin{table*}[t]
\centering
\resizebox{\textwidth}{!}{
\begin{tabular}{@{}l c c c c c@{}}  % @{} removes extra space at the ends of the table
\toprule
\textbf{Method} & \textbf{Human-Computer Interaction} & \textbf{LLM} & \textbf{Required Information} & \textbf{Return Information} & \textbf{Relevance Source} \\
\midrule
ComLittee~\citep{2302HyeonsuB.KangComLittee} & \checkmark & - & Authorship Graphs & Meta data with relevant authors & R, Co, Ci \\
ArZiGo~\citep{2410IratxePinedoArZiGo} & \checkmark & - & User Interest & Meta data & R \\
PaperWeaver~\citep{2403YoonjooLeePaperWeaver} & \checkmark & \checkmark & Collected Papers & Meta data with description & R \\
~\citet{22HyeonsuBKangFrom} & - & - & \makecell{Author's social network relationships \\ +Reference relationship} & Meta data with relevant authors & R \\
\bottomrule
\end{tabular}
}
\caption{Research Paper Recommendation, we referred to ~\citet{2302HyeonsuB.KangComLittee} for comparing different methods, where R represents Paper recommender score, Co represents Co-author relationship, and Ci represents Cited author relationship.}
\label{tab:research_paper_recommendation}
\end{table*}

\begin{table*}[t]
\centering
\resizebox{\textwidth}{!}{ 
\begin{tabular}{@{}l c c c c c c l@{}}
\toprule
\textbf{Method} & \textbf{Research Field} & \textbf{Across Stages} & \textbf{Human Interaction} & \textbf{Task} & \textbf{Input} & \textbf{Output} & \textbf{Evaluation Method} \\
\midrule
AutoSurvey~\citep{2406YidongWangAutoSurvey} & Any & \checkmark & - & \makecell{Outline Generation, \\ +Full-text Generation} & Title \& Full Content & Literature Survey & LLM \& Human \\
CHIME~\citep{2407ChaoChunHsuCHIME} & Biomedicine & - & \checkmark & Outline Generation & Title \& Full Content & Hierarchical Outline & Automatic Metrics \\
Knowledge Navigator~\citep{2408UriKatzKnowledge} & Any & - & - & Outline Generation & Title \& Full Content & Hierarchical Outline & LLM \& Human \\
Relatedly~\citep{PalaniNDZBC23} & Any & - & - & Full-text Generation & Title \& Related Work & Literature Survey & Human \\
STORM~\citep{2402YijiaShaoAssisting} & Any & - & - & \makecell{Outline Generation, \\ +Full-text Generation} & Title \& Full Content & Literature Survey & LLM \& Automatic Metrics \\
\bottomrule
\end{tabular}
}
\caption{Scientific Literature Review, we referred to \citet{2402FranciscoBolaArtificial} and made modifications, thereby comparing different methods.}
\label{tab:scientific_lit_review}
\end{table*}

\begin{table*}[t]
\centering
\resizebox{\textwidth}{!}{ 
\begin{tabular}{@{}l c c c c c c c c c@{}}
\toprule
\textbf{Method} & \textbf{Research Field} & \textbf{Across Stages} & \textbf{Human Interaction} & \textbf{Multi-agent} & \textbf{Trained Model} & \textbf{Online RAG} & \textbf{Novelty} & \textbf{Feasibility} & \textbf{Validity} \\
\midrule
COI~\citep{2410LongLiChain} & Any & \checkmark & - & - & - & \checkmark & \checkmark & \checkmark & \checkmark \\
Learn2Gen~\citep{2412RuochenLiLearning} & Artification Intelligence & \checkmark & - & - & \checkmark & - & \checkmark & \checkmark & \checkmark \\
MatPilot~\citep{2411ZiqiNiMatPilot} & Materials Science & \checkmark & \checkmark & \checkmark & - & - & \checkmark & \checkmark & - \\
SciAgents~\citep{2409AlirezaGhafarollahiSciAgents} & Any & - & \checkmark & \checkmark & - & - & \checkmark & \checkmark & - \\
SciMON~\citep{2305QingyunWangSciMON} & Any & - & - & - & \checkmark & - & \checkmark & - & - \\
\bottomrule
\end{tabular}
}
\caption{Hypothesis Generation, we referred to \citet{25LuoZimingLLM4SR} and made modifications, thereby comparing different methods.}
\label{tab:hypothesis_generation}
\end{table*}

\begin{table*}[t]
\centering
\resizebox{\textwidth}{!}{ 
\begin{tabular}{@{}l c c c c c c l@{}}
\toprule
\textbf{Method} & \textbf{Input} & \textbf{Document Retrieval} & \textbf{Human Interaction} & \textbf{Rationale Selection} & \textbf{Evidence Format} & \textbf{Output} \\
\midrule
MULTIVERS~\citep{2112DavidWaddenMultiVerS} & Claim \& scientific abstract & Provided & - & Longformer & Document & Label \& sentence-level rationales \\
SFAVEL~\citep{2309AdrinBazagaUnsupervised} & Claim & Pre-trained Language Model & - & - & knowledge graph & Top-K Facts \& Corresponding Relevance Scores \\
ProToCo~\citep{2306FengzhuZengPrompt} & Claim-Evidence Pair & Provided & - & - & Sentence & Label \\
MAGIC~\citep{24WeiYuKaoMAGIC} & Claim & Provided & - & Dense Passage Retriever & Sentence & Label \\
aedFaCT~\citep{2305EnesAltuncuaedFaCT} & News Article & Google Search & \checkmark & Human & Document & Evidence \\
\bottomrule
\end{tabular}
}
\caption{Scientific Claim Verification, we referred to \citet{2305JurajVladikaScientific} and made modifications, thereby comparing different methods.}
\label{tab:scientific_claim_verification}
\end{table*}

\begin{table*}[t]
\centering
\resizebox{\textwidth}{!}{ 
\begin{tabular}{@{}l c c c c c@{}}
\toprule
\textbf{Method} & \textbf{Generation Based} & \textbf{Stepwise} & \textbf{Heuristic Search} & \textbf{Informal or Formal} & \textbf{Human-authored Realistic Proof} \\
\midrule
IBR~\citep{QuCGDX22} & - & \checkmark & \checkmark & Informal & - \\
GPT-f~\citep{2009StanislasPoluGenerative} & \checkmark & \checkmark & - & Formal & \checkmark \\
DT-Solver~\citep{23DT-SolverHaimingWang} & \checkmark & \checkmark & \checkmark & Formal & \checkmark \\
POETRY~\citep{2405HaimingWangProving} & \checkmark & - & - & Formal & \checkmark \\
\bottomrule
\end{tabular}
}
\caption{Theorem proving, we referred to \citet{2303ZonglinYangLogical} and made modifications, thereby comparing different methods.}
\label{tab:theorem_proving}
\end{table*}

\begin{table*}[t]
\centering
\resizebox{\textwidth}{!}{ 
\begin{tabular}{@{}l c c c c l l c@{}}
\toprule
\textbf{Method} & \textbf{Research Field} & \textbf{Across Stages} & \textbf{Human Interaction} & \textbf{Multi-agent} & \textbf{Task} & \textbf{Input} & \textbf{External tools} \\
\midrule
AutoML-GPT~\citep{2305ShujianZhangAutoMLGPT} & Artification Intelligence & - & - & - & Automated Machine Learning & Task-oriented Prompts & - \\
Chemcrow~\citep{2304AndresMBranAugmenting} & Chemistry & - & \checkmark & - & Chemical Task & Task Description & \checkmark \\
DOLPHIN~\citep{2501YuanJiakangDolphin} & Any & \checkmark & - & \checkmark & Automated Scientific Research & Idea & \checkmark \\
MechAgents~\citep{2311BoNiMechAgents} & Physics & - & - & \checkmark & Mechanical Problem & - & - \\
~\citet{2404ManningBenjaminSAutomated} & Social Science & \checkmark & - & \checkmark & Simulating Human & - & - \\
\bottomrule
\end{tabular}
}
\caption{Experiment Validation: we use attribute diagrams to compare different schemes, and the table design refers to Hypothesis Generation.}
\label{tab:experiment_validation}
\end{table*}

\begin{table*}[t]
\centering
\resizebox{\textwidth}{!}{ 
\begin{tabular}{@{}l c c l l l@{}}
\toprule
\textbf{Method} & \textbf{Across Stages} & \textbf{Human Interaction} & \textbf{Task} & \textbf{Input} & \textbf{Evaluation Method} \\
\midrule
AI Scientist~\citep{2408ChrisLuThe} & \checkmark & - & Full-text Generation & Manuscript Template \& Experimental Results \& Hypothesis & LLM \\
data-to-paper~\citep{2404TalIfarganAutonomous} & \checkmark & \checkmark & Full-text Generation & Experimental Results \& Hypothesis & - \\
ScholaCite~\citep{2402MartinBoyleAnnaShallow} & - & - & Related Work Generation & Title \& Abstract \& Citation & Citation Graph Metrics \\
SciLit~\citep{2306NianlongGuSciLit} & \checkmark & - & Citation Generation & Keywords & Automatic Metrics \\
~\citet{24GuNianlongControllable} & - & - & Citation Generation & Citation Intent \& Keywords & Human \\
\bottomrule
\end{tabular}
}
\caption{Manuscript Writing, we referred to \citet{2201XiangciLiAutomatic, 2404XiangciLiRelated} and made modifications, thereby comparing different methods.}
\label{tab:manuscript_writing}
\end{table*}

\begin{table*}[t]
\centering
\resizebox{\textwidth}{!}{%
\begin{tabular}{@{}lccccccl@{}}
\toprule
\textbf{Method} & \textbf{Across Stages} & \textbf{Human Interaction} & \textbf{Paper Review} & \textbf{Meta Review} & \textbf{Multi-agent} & \textbf{Trained Model} & \textbf{Output} \\
\midrule
Gamma-Trans~\citep{22PanitanMuangkammuenExploiting} & - & - & \checkmark & - & - & \checkmark & Peer-review Score \\
MARG~\citep{2401MikeDArcyMARG} & - & - & \checkmark & - & \checkmark & - & Peer-review Comments \\
CycleResearcher~\citep{2411YixuanWengCycleResearcher} & \checkmark & - & \checkmark & - & - & \checkmark & Peer-review Comments \& Score \\
PeerArg~\citep{2409PurinSukpanichnantPeerArg} & - & - & - & \checkmark & - & - & Final Decision \\
GLIMPSE~\citep{2406MaximeDarrinGLIMPSE} & - & \checkmark & - & \checkmark & - & - & Summary of Peer-review \\
\bottomrule
\end{tabular}
}
\caption{Peer Review, we referred to \citet{2111JialiangLinAutomated} and made modifications, thereby comparing different methods.}
\label{tab:peer_review}
\end{table*}

\begin{figure*}[t]
    % \vspace{-1cm}
    \centering
    \resizebox{\linewidth}{!}{
        \begin{forest}
            forked edges,
            for tree={
                child anchor=west,
                parent anchor=east,
                grow'=east,
                anchor=west,
                base=left,
                font=\large, 
                rectangle,
                draw=hidden-black,
                rounded corners,
                minimum height=2.5em, 
                minimum width=5em,    
                edge+={darkgray, line width=1pt},
                s sep=6pt,          
                inner xsep=0.6em,    
                inner ysep=0.9em,     
                line width=0.8pt,
                text width=10em,     
                ver/.style={
                    rotate=90,
                    child anchor=north,
                    parent anchor=south,
                    anchor=center,
                    text width=13em   
                },
                leaf/.style={
                    text opacity=1,
                    fill opacity=.5,
                    fill=hidden-blue!90, 
                    text=black,
                    text width=50em, 
                    font=\large,
                    inner xsep=0.6em,
                    inner ysep=0.9em,
                    draw,
                }, 
            },
            [
                A survey of AI for Research, ver
                [
                    Hypothesis \\Formulation ~(\S\ref{sec:HypothesisFormulation})
                    [
                        Knowledge Synthesize ~(\S\ref{sec:HypothesisFormulation_KnowledgeSynthesis})
                        [
                            Research Paper \\ Recommendation ~(\S\ref{sec:HypothesisFormulation_KnowledgeSynthesis_RPR})
                            [   
                                PaperWeaver~\cite{2403YoonjooLeePaperWeaver}{,}
                                ArZiGo~\cite{2410IratxePinedoArZiGo}{,}
                                ComLittee~\cite{2302HyeonsuB.KangComLittee}{,}
                                ~\citet{24VaiosStergiopoulosAn}{,}
                                ~\citet{22HyeonsuBKangFrom}{,}
                                ~\citet{2201ChristinKatharinaKreutzScientific} {,}
                                ~\citet{20shahidinsights}{,}
                                ~\citet{19XiaomeiBaiScientific}{,}
                                ~\citet{1901ZhiLiA}{,}
                                ~\citet{16beelpaper}
                                , leaf, text width=50em
                            ]
                        ]
                        [
                            Systematic \\Literature Review ~(\S\ref{sec:HypothesisFormulation_KnowledgeSynthesis_SLR})
                            [
                                AutoSurvey ~\cite{2406YidongWangAutoSurvey} {,}
                                OpenScholar ~\cite{2411AkariAsaiOpenScholar} {,}
                                HiReview ~\cite{2410HuYuntongHireview}{,}
                                PaperQA2 ~\cite{2409MichaelDSkarlinskiLanguage}{,}
                                Knowledge Navigator ~\cite{2408UriKatzKnowledge}{,}
                                CHIME ~\cite{2407ChaoChunHsuCHIME}{,}
                                STORM ~\cite{2402YijiaShaoAssisting}{,}
                                LitLLM ~\cite{2402ShubhamAgarwalLitLLM}{,}
                                Bio-SIEVE ~\cite{2308AmbroseRobinsonBio-SIEVE}{,}
                                KGSum ~\cite{2209PanchengWangMulti-Document}{,}
                                \citet{2412agarwalllms}{,}
                                \citet{2408YuxuanLaiInstruct}{,}
                                \citet{2404Teosusnjakautomating}{,}
                                \citet{2402FranciscoBolaArtificial}{,}
                                \citet{24BlockWhat}{,}
                                \citet{2304KunZhuHierarchical} {,}
                                \citet{22NoufIbrahimAltmamiAutomatic}
                                , leaf, text width=50em
                            ]
                        ]
                    ]
                    [
                        Hypotheses Generation ~(\S\ref{sec:HypothesisFormulation_HypothesesGeneration})
                        [
                            Other Works
                            [
                                MOOSE-Chem ~\cite{2410ZonglinYangMOOSE}{,}
                                IdeaSynth ~\cite{2410KevinPuIdeaSynth}{,}
                                \citet{2409ChengleiSiCan}{,}
                                \citet{2409SandeepKumarCan}{,}
                                \citet{23GirotraIdeas}{,}
                                \citet{2311BiQingQiLarge}{,}
                                \citet{2210MarioKrennPredicting}{,}
                                \citet{1708SamHenryLiterature}
                                , leaf, text width=50em
                            ]
                        ]
                        [
                            Input Data Quality
                            [
                                SciPIP~\cite{2410WenxiaoWangSciPIP}{,}
                                COI~\cite{2410LongLiChain}{,}
                                SciAgents~\cite{2409AlirezaGhafarollahiSciAgents}{,}
                                DATAVOYAGER ~\cite{2402BodhisattwaPrasadMajumder}{,}
                                SciMON~\cite{2305QingyunWangSciMON}{,}
                                \citet{2403MarkusJBuehlerAccelerating}{,}
                                \citet{2410HaokunLiuLiterature}
                                , leaf, text width=50em
                            ]
                        ]
                        [
                            Hypothesis Quality
                            [
                                Agentrxiv~\cite{schmidgall2025agentrxiv}
                                Dolphin~\cite{2501YuanJiakangDolphin}{,}
                                VIRSCI~\cite{2410HaoyangSuTwo}{,}
                                Nova~\cite{2410XiangHuNova}{,}
                                AI Scientist~\cite{2408ChrisLuThe}{,}
                                SGA~\cite{2405PingchuanMaLLM}{,}
                                HypoGeniC ~\cite{2404YangqiaoyuZhouHypothesis}{,}
                                ResearchAgent~\cite{2404JinheonBaekResearchAgent}{,}
                                Acceleron~\cite{2403HarshitNigamAcceleron}{,}
                                MOOSE ~\cite{2309ZonglinYangLarge}
                                , leaf, text width=50em
                            ]
                        ]
                    ]
                ]
                [
                    Hypothesis Validation  ~(\S\ref{sec:HypothesisValidation})
                    [
                        Scientific Claim Verification ~(\S\ref{sec:HypothesisValidation_ScientificClaimVerification})
                        [
                            Claim
                            [
                                HiSS~\cite{2310XuanZhangTowards}{,}
                                SFAVEL~\cite{2309AdrinBazagaUnsupervised}{,}
                                ProToCo~\cite{2306FengzhuZengPrompt}{,}
                                \citet{2402AmelieWhrlWhat}
                                , leaf, text width=50em
                            ]
                        ]
                        [
                            Evidence
                            [
                                \citet{2408MaxGlocknerGrounding}{,}
                                \citet{2404JurajVladikaImproving}{,}
                                \citet{2402JurajVladikaComparing}{,}
                                \citet{2402AmelieWhrlUnderstanding}{,}
                                \citet{2309LiangmingPanInvestigating}{,}
                                \citet{2210MaxGlocknerMissing}{,}
                                \citet{2112DavidWaddenMultiVerS}
                                , leaf, text width=50em
                            ]
                        ]
                        [
                            Verification
                            [
                                ClaimVer~\cite{2403PreetamPrabhuSrikarDammuClaimVer}{,}
                                MAGIC~\cite{24WeiYuKaoMAGIC}{,}
                                FactKG~\cite{2305JihoKimFactKG}{,}
                                aedFaCT~\cite{2305EnesAltuncuaedFaCT}{,}
                                PROGRAMFC~\cite{2305LiangmingPanFact-Checking}{,}
                                \citet{2410XiaochengZhangAugmenting}{,}
                                \citet{2409AriefPurnamaMuharramEnhancing}{,}
                                \citet{2407IslamEldifrawiAutomated}{,}
                                \citet{2402YupengCaoCan}{,}
                                \citet{2301AnubrataDasThe}{,}
                                \citet{23JinxuanWuCharacterizing}
                                , leaf, text width=50em
                            ]
                        ]
                    ]
                    [
                        Theorem Proving ~(\S\ref{sec:HypothesisValidation_TheoremProving})
                        [
                            DeepSeek-Prover~\cite{2410HuajianXinDeepSeek-Prover}{,}
                            Lean-STaR~\cite{2407HaohanLinLean-STaR}{,}
                            Thor~\cite{2407AlbertQiaochuJiangAdvances}{,}
                            COPRA~\cite{24ThakurAn}{,}
                            Logo-power~\cite{2310HaimingWangLEGO-Prover}{,}
                            Baldur~\cite{2303EmilyFirstBaldur}{,}
                            Mustard~\cite{2303YinyaHuangMUSTARD}{,}
                            DT-Solver~\cite{23DT-SolverHaimingWang}{,}
                            HTPS~\cite{2205GuillaumeLampleHyperTree}{,}
                            GPT-f~\cite{2009StanislasPoluGenerative}{,}
                            \citet{2405HaimingWangProving}{,}
                            \citet{2404PeiyangSongTowards}{,}
                            \citet{2404ZhaoyuLiA}{,}
                            \citet{2305XueliangZhaoDecomposing}{,}
                            \citet{2303ZonglinYangLogical}{,}
                            \citet{2210AlbertQiaochuJiangThe}
                            , leaf, text width=62.5em
                        ]
                    ]
                    [
                        Experiment Validation ~(\S\ref{sec:HypothesisValidation_ExperimentVerification})
                        [
                            MatPilot~\cite{2411ZiqiNiMatPilot}{,}
                            CRISPR-GPT~\cite{2404KaixuanHuangCRISPRGPT}{,}
                            AgentHPO ~\cite{2402SiyiLiuLarge}{,}
                            AutoML-GPT~\cite{2305ShujianZhangAutoMLGPT}{,}
                            MLCopilot~\cite{2304LeiZhangMLCopilot}{,}
                            \cite{2406SrenArltMetaDesigning}{,}
                            \citet{2402SubbaraoKambhampatiPosition}{,}
                            \citet{24RuanYixiangAn}{,}
                            \citet{2310QianHuangMLAgentBench}{,}
                            \citet{23DaniilABoikoAutonomous}{,}
                            \citet{2304AndresMBranAugmenting}{,}
                            \citet{23SzymanskiNathanJautonomousAn}{,}
                            \citet{2305RuiboLiuTraining}{,}
                            \citet{2404ManningBenjaminSAutomated}{,}
                            \citet{2412XinyiMouFrom}
                            , leaf, text width=62.5em
                        ]
                    ]
                ]
                [
                    Manuscript\\ Publication ~(\S\ref{sec:ManuscriptPublication})
                    [
                        Manuscript Writing ~(\S\ref{sec:ManuscriptPublication_ManuscriptWriting})
                        [
                            Citation Text \\Generation
                            [
                                SciLit~\cite{2306NianlongGuSciLit}{,}
                                DisenCite~\cite{2206YifanWangDisenCite}{,}
                                \citet{24GuNianlongControllable}{,}
                                \citet{2305TianyuGaoEnabling}{,}
                                \citet{22YuMengxiaScientific}{,}
                                \citet{22JungShingYunIntent}
                                , leaf, text width=50em
                            ]
                        ]
                        [
                            Related Work \\Generation
                            [
                                ScholaCite~\cite{2402MartinBoyleAnnaShallow}{,}
                                UR3WG~\cite{23ZhengliangShiTowards}{,}
                                \citet{2404XiangciLiRelated}{,}
                                \citet{24LuyaoYuReinforced}{,}
                                \citet{24NishimuraToward}{,}
                                \citet{2201XiangciLiAutomatic}
                                , leaf, text width=50em
                            ]
                        ]
                        [
                            Complete Manuscripts \\Generation
                            [
                                Cocoa~\cite{2412KJKevinFengCocoa}{,}
                                Step-Back~\cite{2406XiangruTangStep}{,}
                                data-to-paper~\cite{2404TalIfarganAutonomous}{,}
                                ARIES~\cite{2306MikeDArcyARIES}{,}
                                OREO~\cite{2204JingjingLiText}{,}
                                R3~\cite{2204WanyuDuRead}{,}
                                \citet{2408YuxuanLaiInstruct}{,}
                                \citet{2405EricChamounAutomated}{,}
                                \citet{24MiltonPividoriA}{,}
                                \citet{2303LaneJourdanText}{,}
                                \citet{2310LinZhichengTechniques}
                                , leaf, text width=50em
                            ]
                        ]
                    ]
                    [
                        Peer Review ~(\S\ref{sec:ManuscriptPublication_PeerReview})
                        [
                            Paper Review \\Generation
                            [
                                CycleResearcher~\cite{2411YixuanWengCycleResearcher}{,}
                                OpenReviewer~\cite{2412MaximilianIdahlOpenReviewer}
                                RelevAI-Reviewer~\cite{2406PauloHenriqueCoutoRelevAI}{,}
                                AgentReview~\cite{2406YiqiaoJinAgentReview}{,}
                                SWIF2T~\cite{2405EricChamounAutomated}{,}
                                MARG~\cite{2401MikeDArcyMARG}{,}
                                Quality Assist~\cite{22BasukiSetioThe}{,}
                                KID-Review~\cite{22WeizheYuanKID}
                                \citet{2406ChengTanPeer}{,}
                                \citet{2310WeixinLiangCan}{,}
                                \citet{2307ZacharyRobertsonGPT4}{,}
                                \citet{22PanitanMuangkammuenExploiting}{,}
                                , leaf, text width=50em
                            ] 
                        ]
                        [
                            Meta-Review \\Generation
                            [
                                Deeprview~\cite{zhu2025deepreview}
                                PeerArg~\cite{2409PurinSukpanichnantPeerArg}
                                GLIMPSE~\cite{2406MaximeDarrinGLIMPSE}{,}
                                MetaWriter~\cite{24LuSunMetaWriter}{,}
                                CGI2~\cite{2305QiZengMeta}{,}
                                \citet{2402MiaoLiA}{,}
                                \citet{2402ShubhraKantiKarmakerSantuPrompting}{,}
                                \citet{2310SandeepKumarWhen}{,}
                                \citet{2305MiaoLiSummarizing}
                                , leaf, text width=50em
                            ]
                        ]
                    ]
                ]
            ]
        \end{forest}
    }
    % \vspace{-6mm}
    \caption{Taxonomy of Hypothesis
Formulation, Hypothesis Validation and Manuscript
Publication (Full Edition).}
    \label{fig:taxonomy_full}
    % \vspace{-3mm}
\end{figure*}
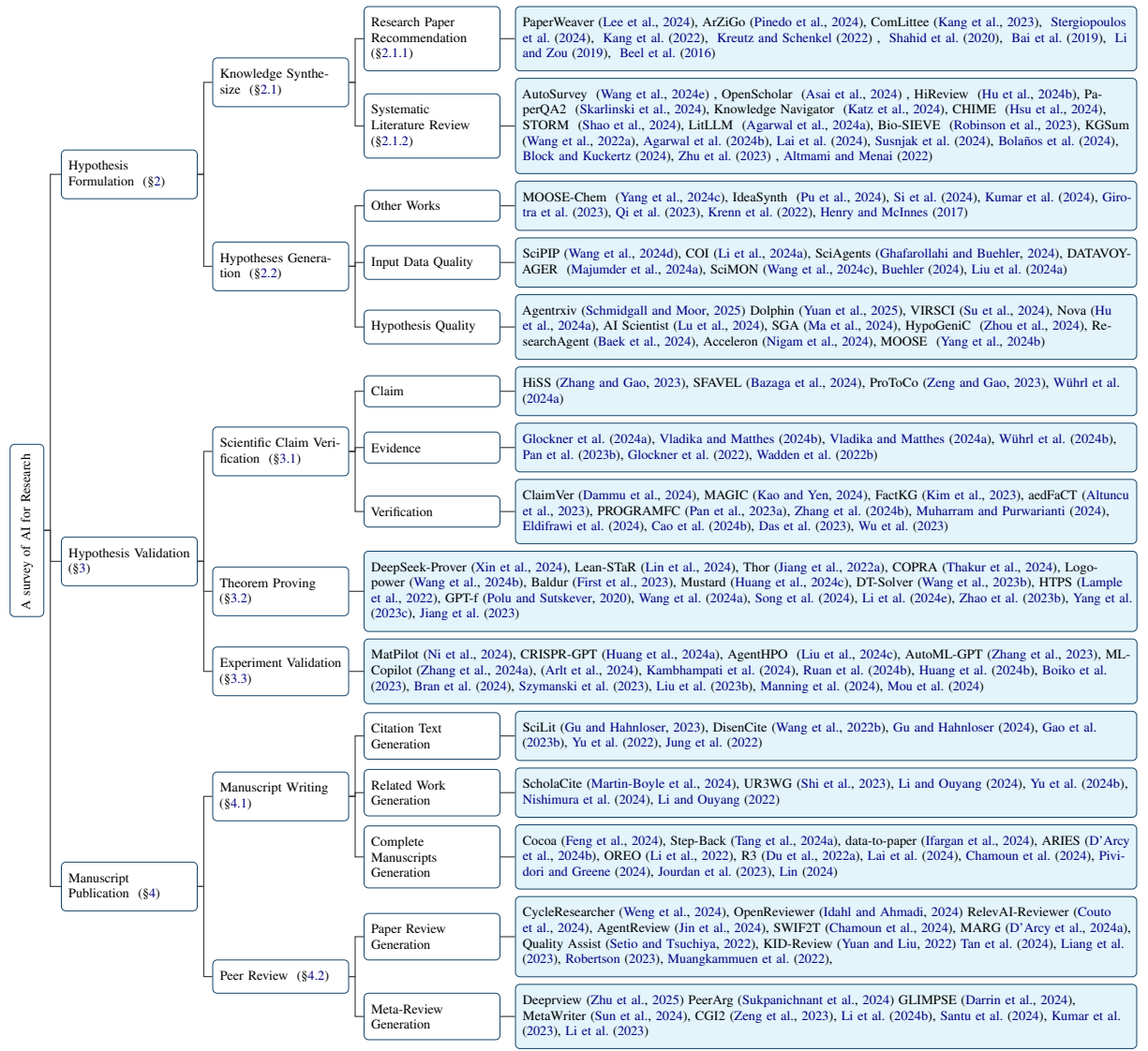

\end{document}